\documentclass[10pt,twocolumn,letterpaper]{article}
\usepackage[T1]{fontenc}  %

\usepackage{cvpr}      %

\DeclareRobustCommand{\method}{%
  \texorpdfstring{E\textsuperscript{2}Forecast}{E2Forecast}%
}

\usepackage{multirow}
\usepackage{tikz}
\usetikzlibrary{arrows.meta,positioning,calc,fit,backgrounds,shapes.geometric,%
                decorations.pathreplacing,decorations.pathmorphing,patterns}

\definecolor{loccol}{RGB}{ 55,105,175}   %
\definecolor{obscol}{RGB}{ 30,140,110}   %
\definecolor{tgtcol}{RGB}{215,125, 30}   %
\definecolor{flwcol}{RGB}{140, 85,175}   %
\definecolor{datecol}{RGB}{200, 85,125}   %
\definecolor{gaucol}{RGB}{ 20,130,165}   %
\definecolor{gryc}{RGB}{120,120,120}

\usepackage{placeins}

\definecolor{cvprblue}{rgb}{0.21,0.49,0.74}
\usepackage[pagebackref,breaklinks,colorlinks,allcolors=cvprblue]{hyperref}

\def\paperID{*****} %
\def\confName{CVPR}
\def\confYear{2027}

\title{Forecasting Earth Surface Change in Embedding Space}

\author{
Jingtao Li$^{1}$\quad
Sadiq Jaffer$^{1}$\quad
Anil Madhavapeddy$^{1}$\quad
Pedro Sousa$^{1}$\quad
Anthony Ortiz$^{2}$\\
Kyle Gao$^{3}$\quad
Zhengpeng Feng$^{1}$\quad
Yanfei Zhong$^{4}$\quad
Kevin White$^{2}$\quad
Srinivasan Keshav$^{1}$\\[4pt]
$^{1}$Department of Computer Science, University of Cambridge\quad
$^{2}$Microsoft AI for Good Research Lab\\
$^{3}$Aalto University\quad
$^{4}$Wuhan University\\[2pt]
{\tt\small \{jl2692, sj514, avsm2, pmms2, zf281, sk818\}@cam.ac.uk},\\
{\tt\small \{anthony.ortiz,kevin.white\}@microsoft.com},\\
{\tt\small kyle.gao@aalto.fi},\quad
{\tt\small zhongyanfei@whu.edu.cn}
}

\begin{document}
\maketitle
\begin{abstract}
Repeated satellite imagery permits understanding of how our planet changes over time,
letting us predict its future and plan mitigating actions.
Most existing studies that rely on a time series of satellite
observations forecast the spectral values at each pixel for a future target date.
However, clouds, shadows, and sensor noise often corrupt these observations, so that
spectral forecasters waste much of their capacity on low-level features or noise rather than real change.
With the advent of Earth foundation models, we argue that forecasting in \textit{embedding space} is a feasible and more effective alternative.
To this end, we design a token space that handles irregular histories from different sensors and foundation models and supports forecasts at any query date.
Tokens from this space are inputs to our proposed forecaster, \method{}, which is a transformer with two branches: a Gaussian branch that directly predicts a mean point forecast and per-dimension standard deviations, and a rectified-flow branch that generates forecast ensembles.
On Sentinel-1 and Sentinel-2 benchmarks, forecasting embeddings instead of spectral values halves the cosine distance (from 0.20 to 0.10) and greatly reduces overconfidence, lowering the z-score standard deviation from 2.63 to 1.22.
By comparing new observations with its forecasts, \method{} can also detect anomalous changes \textit{without} labels or retraining, raising F$_1$ from 41.58 to 48.70 on a global benchmark. This consistent superiority is validated on nine remote sensing foundation models. 
Beyond change detection, 
we believe our approach advances the field by permitting effective 
reuse of diverse pretrained task heads. 
Our code is publicly available at https://github.com/ucam-eo/e2forecast.
\end{abstract}    
\section{Introduction}
\label{sec:intro}

\begin{figure*}[t]
  \centering
  \includegraphics[width=\textwidth]{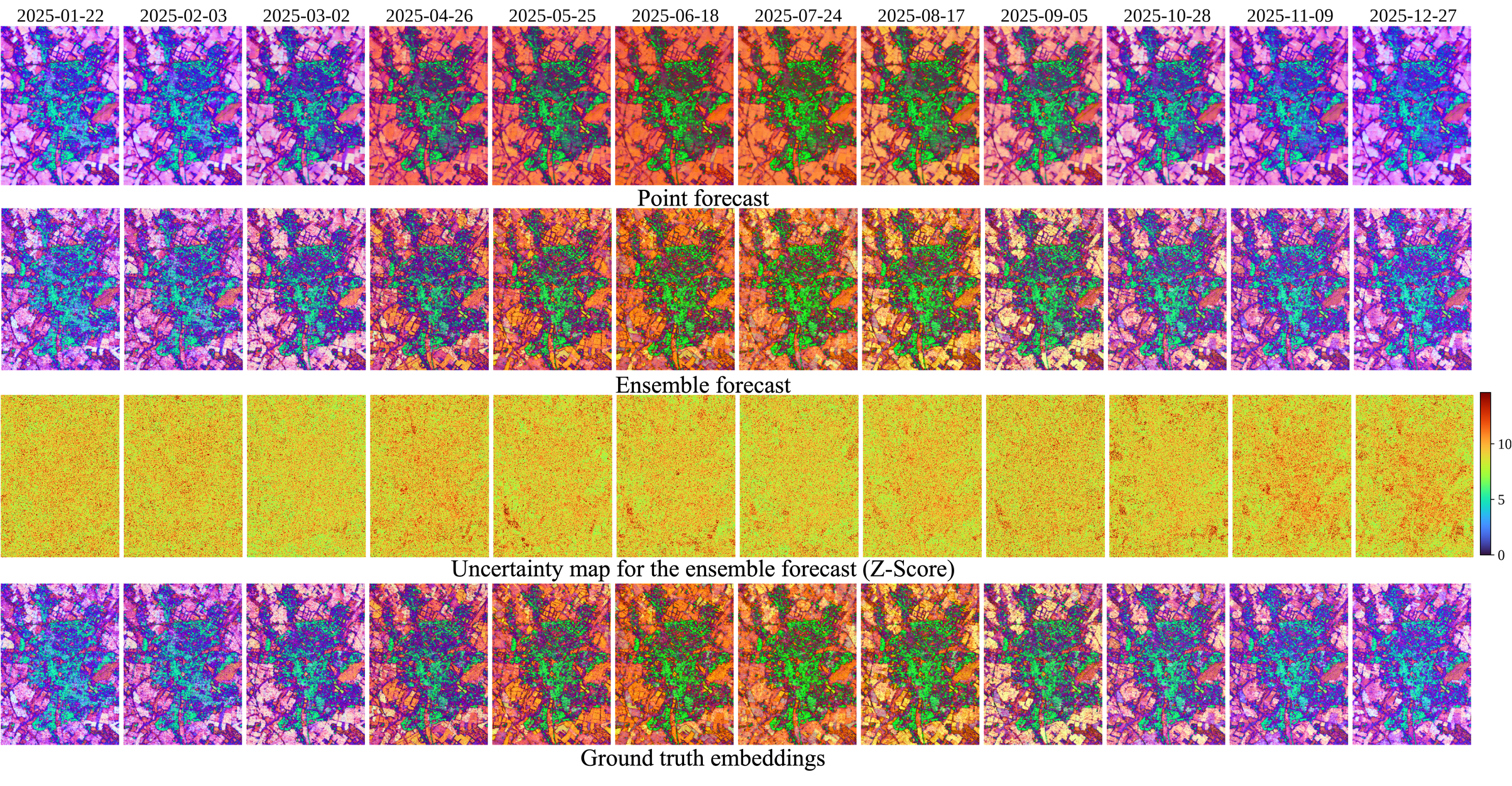}
  \caption{\textbf{A demo of \method{}'s flexible forecasting with any query date, producing both point and ensemble forecasts with uncertainty.}
  We choose this Cambridge scene because it includes both urban and rural areas.
  With time-series OlmoEarth embeddings, we train \method{} on 2024 only and use January 2025 as context, so every later forecast comes from learned temporal patterns rather than copied observations.
  Both point and ensemble forecasts of 10 generated embeddings capture the seasonal changes effectively, while ensemble forecasts can better describe the diversity and details.}
  \label{fig:cambridge_example}
\end{figure*}

The Earth's surface is dynamic, where 
crops grow and are harvested, forests burn and regenerate, cities expand, and populations move.
Most of these changes follow complex, seasonal, and long-term patterns.
Understanding these patterns allows us to forecast the future and prepare in advance, 
which is valuable for food security, disaster response, and environmental management.

The remote sensing community has become increasingly proficient in forecasting future satellite images~\cite{requenamesa2021earthnet,zhang2025units,gu2025long} and
physical variables~\cite{bi2023pangu}. However, most studies model both the dynamics and the forecast in 
the original signal (spectral) space.
Consequently, the model has to deal with artifacts such as clouds, shadows, and sensor noise,
which vary between acquisitions
but say little about how the land changes~\cite{stewart2026earthembeddings}. 
We demonstrate (\S\ref{sec:contributions}) that signal-space forecasts lead models to miss many high-level features. 
Moreover, due to significant differences in sensor modalities, such as SAR and optical, 
it is difficult to forecast flexible multi-modal observations.

In recent years, geospatial foundation models have been used to create \textit{embeddings}
that turn noisy and 
heterogeneous observations into compact and semantically meaningful vectors~\cite{brown2025alphaearth,feng2026tessera,feng2026tesserav2,herzog2025olmoearth,szwarcman2024prithvi}.
These embeddings suppress much of the acquisition-specific variation while preserving the
semantic content of the observations. 
Their wide impact can be witnessed in diverse downstream tasks such as crop mapping, landslide detection, and habitat analysis ~\cite{ma2026harvesting,cheng2026landslide,ryan2026wetland}. 
Existing work uses embeddings to describe what has already been observed, but does not explicitly model their evolution to forecast future states.
Instead, we seek to \textit{directly forecast in  the embedding space} and
exploit different foundation models that can generate embeddings for different sensors, spatial resolution, patch size, and time durations. 

Unlike image-space forecasting, which predicts a small number of spectral channels per pixel, embedding-space forecasting predicts a high-dimensional feature vector for each token. This poses two main challenges:
(i) \textit{How can we design a token space that supports forecasts from irregularly timed inputs?}
Satellite observations are rarely regular and revisit intervals differ between sensors and clouds can remove 
blocks of observations, so two consecutive embeddings may be days or
months apart. Most forecasting models assume 
input observations are spaced by regular steps and predict assuming a fixed lead time ~\cite{liu2026semantic, liu2025timerxl}. 
(ii) \textit{How can we model multiple possible outputs?}
Even with a complete embedding history,
the future, by definition, remains uncertain. For example, the croplands can be either rotated or left fallow, or a seashore may or may not have vessels. 
A forecast should ideally
generate an ensemble of potential futures, each with a corresponding uncertainty. Existing models (\S\ref{sec:related}) only partly meet this requirement.
For example, probability
forecasting models give the most likely future distribution and its uncertainty,
but it is difficult to draw realistic, potentially correlated features from high-dimensional distributions \cite{salinas2020deepar,lim2021temporal}. 
Some generative models can produce realistic ensemble
forecasts~\cite{kollovieh2023predict,li2024transformer,ding2026simdiff}, but
their most likely future and uncertainty are only known after drawing at least 100 samples with tens of denoising steps, which makes it inefficient.

\subsection{Forecasting in embedding space}
\label{sec:contributions}

We propose \method{}, which forecasts Earth embeddings at arbitrary lead times from irregularly sampled historical embeddings.

It can give both the Gaussian mean as a point forecast and ensemble forecasts (\cref{fig:overview}), alongside the per-dimensional standard deviation to measure the uncertainty. 
To avoid  assuming a fixed time step, we design a \textit{forecasting token space} that turns
each historical embedding into a token that carries its own acquisition date, and
adds an adjustable query date to control the lead time.  
Using these tokens, we train a single transformer with two branches to produce Gaussian probability estimates and generated ensemble embeddings. Gaussian branch predicts a Gaussian mean, used as the point forecast, alongside the standard deviation directly estimated to measure the uncertainty. The embedding generation branch learns a rectified flow and draws
realistic ensemble forecasts with different noise initializations. 
An example in \Cref{fig:cambridge_example} shows the role of each branch and highlights the model's flexible forecasting capability. 
With the forecasted embeddings, users can either compare them against observed embeddings to detect anomalies (shown later), or reuse the diverse pre-trained task heads to perform downstream analysis directly.

By working in embedding space, \method{} is not tied to any one sensor or
foundation model. We have tested 9 different embedding spaces such as OlmoEarth~\cite{herzog2025olmoearth}, Prithvi-EO-2.0~\cite{szwarcman2024prithvi}, AlphaEarth~\cite{brown2025alphaearth},
Tessera~\cite{feng2026tessera} and the SatCLIP vision encoder ~\cite{klemmer2024satclip}, using both the Sentinel-1 and Sentinel-2 time series.
With our approach, we observe consistent and large improvements in both forecast accuracy and uncertainty estimation. 
Remarkably, this improvement advances the SOTA in anomaly change detection \textit{without} any label or head, 
simply by scoring a new observation by its distance to the forecast, thus
raising the anomaly change detection F$_1$ from 41.58 to 48.70 on a global benchmark covering 19{,}801.60\,km$^2$. 

Our contributions are to:
\begin{itemize}
  \item forecast in Earth embedding space to alleviate low-level noise and focus on important information, which is a new paradigm enabled by geospatial foundation models.
  \item design a flexible forecasting model \method{} that accepts diverse sensor embeddings, irregular historical observations, and arbitrary future date queries.
  \item jointly train probability estimation and flow generation in a single transformer, giving an uncertainty-aware point forecast and ensemble forecasts.
  \item perform accurate anomaly change detection directly from the forecasts without retraining the model.
\end{itemize}

\section{Related Work}
\label{sec:related}

\subsection{Earth analysis in embedding space}
\label{sec:rel_embedding}
Earth embeddings compress a time series of repeated Earth observations into a high-level representation space~\cite{stewart2026earthembeddings}. They are produced by geospatial foundation models (GFMs), which are designed for different sensors, resolutions, and time-series settings. For example, CROMA~\cite{fuller2023croma} jointly encodes Sentinel-1 radar and Sentinel-2 optical imagery, DOFA \cite{xiong2024dofa} and HyperFree~\cite{li2025hyperfree} processes images with varied channels, Scale-MAE~\cite{reed2023scalemae} handles varying ground sampling distances, and Prithvi-EO-2.0~\cite{szwarcman2024prithvi}, Galileo~\cite{tseng2025galileo} and OlmoEarth~\cite{herzog2025olmoearth} ingest multi-temporal, multi-sensor inputs. More recently, some embeddings have been made available as precomputed global ready-to-use products~\cite{fang2026products}, including Google's AlphaEarth Foundations~\cite{brown2025alphaearth} and TESSERA~\cite{feng2026tessera} at $10$\,m, ESD~\cite{chen2026esd} at $30$\,m, and Major TOM~\cite{czerkawski2024majortom} at kilometre scale.

Both GFMs and embedding products operationalize the ``embeddings as data'' paradigm by reducing the burden of preprocessing heterogeneous observations and lowering computational and training-data requirements for downstream tasks. This paradigm has been widely adopted for diverse mapping tasks~\cite{ma2026harvesting,cheng2026landslide,ryan2026wetland}. However, current studies primarily exploit the spatial continuity of embeddings; their temporal continuity, which is 
the focus of this work,
remains largely unexplored\cite{stewart2026earthembeddings}.

\subsection{Earth observation forecasting}
\label{sec:rel_eo_prediction}
Forecasting Earth observations has predominantly been for physical variables~\cite{woo2022etsformer,bi2023pangu,sousa2026downscaling} and satellite imagery. Requena-Mesa et al.~\cite{requenamesa2021earthnet} introduced the task in 2021 with the classical EarthNet2021 benchmark, where future Sentinel-2 images are predicted using historical observations and future weather conditions. Earthformer~\cite{gao2022earthformer} then applied cuboid space-time attention to this benchmark, and Contextformer~\cite{benson2024contextformer} improved vegetation forecasting by fusing spatial context with weather drivers. EarthPT~\cite{smith2023earthpt}, instead, autoregressively forecasts pixel-level reflectance time series. More recently, generative models have been introduced to produce realistic future images, including diffusion models conditioned on past acquisitions~\cite{khanna2024diffusionsat}, flow matching-based methods~\cite{zhang2025units}, and video generation-based models~\cite{gu2025long}.

Despite their architectural and application differences, prior work on forecasting is typically in the original signal space, which is inevitably affected by observation-specific variabilities~\cite{stewart2026earthembeddings}, such as clouds, shadows, and sensor noise ~\cite{stewart2026earthembeddings}. We will show that although the forecasting images can look realistic, there are large errors when forecasts are converted into embedding space, implying loss of features.

\subsection{Anomaly change detection}
Not every change on the Earth's surface is an anomaly. Time series forecasting can help identify normal changes such as 
crops growing, seasons turning, and human activity in unsupervised manner. 
For example,  AnomalyCD~\cite{li2026anomalycd} and ESIA~\cite{li2026esia} used long time-series observations to detect abrupt changes, where they assume that changes not seen before are anomalies. Some forecasting-based detection methods identify both gradual and abrupt changes~\cite{verbesselt2010bfast,zhu2014ccdc,pickens2025rapid}.
However, their forecasts are still in the (original) spectral space and only apply to specific sensors.

Since forecasts provide a baseline for expected temporal variation, an accurate forecast can suppress uninteresting normal changes, so that the deviation of new observations from this baseline serves as an anomaly score. From this perspective, we hope that our improvements in forecasting will also benefit anomaly detection.

\section{Earth embedding forecasting}
\label{sec:method}

This section discusses the details of our method, \method{}, which learns the pattern of changes in Earth embedding space, forecasts with a flexible time-series organization, and outputs uncertainty-aware ensemble outputs.
Here we first formalise the embedding forecasting framework (\cref{sec:framework}),
then instantiate it with a novel token space (\cref{sec:backbone}).
We carry out joint
training of probability estimation and embedding generation
(\cref{sec:flow}) and finally describe how we use inferences both for forecasting and for
forecast-based anomaly change detection (\cref{sec:inference}).

\begin{figure*}[t]
  \centering
  \includegraphics[width=\textwidth]{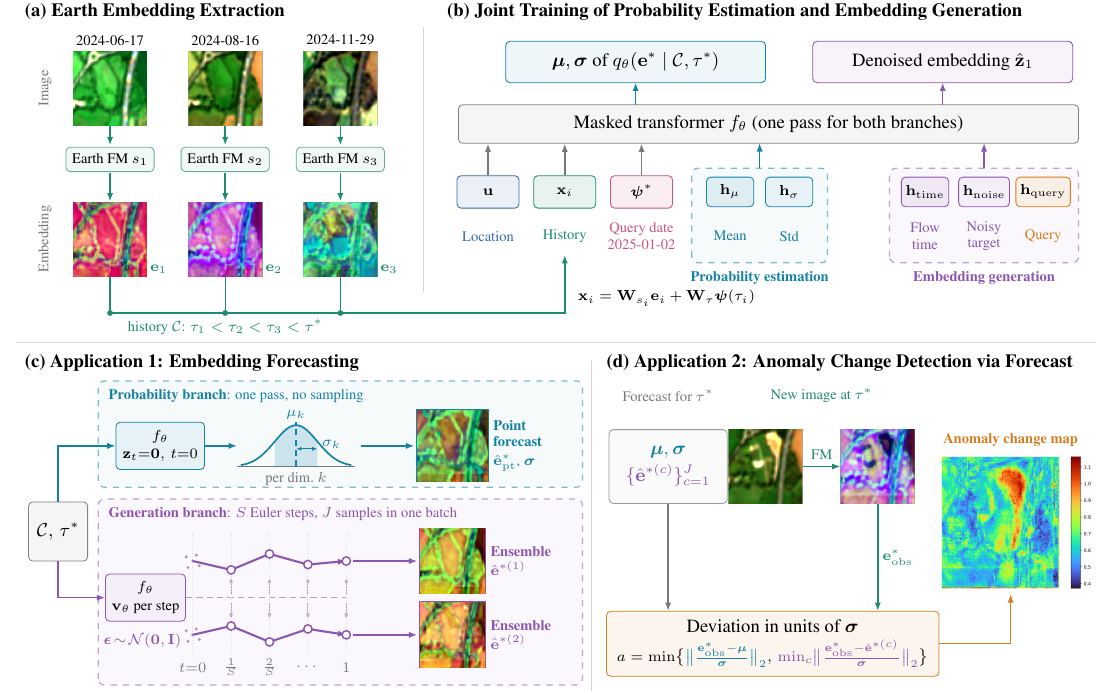}
  \caption{\textbf{Overview of \method{}.}
  (a) Each image in an irregular time series is converted into an Earth
  embedding $\mathbf{e}_i$ by an Earth foundation model $s_i$, which may
  differ across observations.
  (b) The historical embeddings with their dates, the location and the query
  date $\tau^{*}$ form a token sequence for a masked transformer $f_{\theta}$.
  In one pass, the probability-estimation branch predicts the per-dimension
  Gaussian $(\boldsymbol{\mu},\boldsymbol{\sigma})$ of the future embedding,
  and the embedding-generation branch denoises a noisy target
  $\mathbf{z}_t=t\,\mathbf{z}_1+(1-t)\,\boldsymbol{\epsilon}$ by flow matching.
  (c) For forecasting, the probability branch gives a point forecast
  $\hat{\mathbf{e}}^{*}_{\mathrm{pt}}=\boldsymbol{\mu}/\lVert\boldsymbol{\mu}\rVert_2$
  with uncertainty $\boldsymbol{\sigma}$ without sampling, and the generation
  branch turns $J$ noise draws into ensemble forecasts with $S$ Euler steps.
  (d) For anomaly change detection, a new observed embedding is compared with
  both the Gaussian forecast and the closest ensemble forecast in units of
  $\boldsymbol{\sigma}$, giving an anomaly change map without change labels or
  a detection head.}
  \label{fig:overview}
\end{figure*}

\subsection{A flexible ensemble framework for embedding forecasting}
\label{sec:framework}
\method{} operates on embeddings already provided by existing Earth observation foundation models. It is agnostic to the foundation model type, missing values in the time-series, and target forecast date. Formally, let $s_i$ identify the foundation model that produces observation embedding $\mathbf{e}_i\in\mathbb{R}^{d_{s_i}}$. For a fixed imaging location $\ell=(\lambda,\varphi)$, with longitude $\lambda$ and latitude $\varphi$, a forecasting instance consists of a potentially heterogeneous embedding history and a future query date:
\begin{equation}
  \mathcal{C}=\bigl\{(\mathbf{e}_i,s_i,\ell,\tau_i)\bigr\}_{i=1}^{N},
  \qquad
  \tau_1\le\cdots\le\tau_N<\tau^{*},
  \label{eq:instance}
\end{equation}
where $\mathcal{C}$ is the \emph{observation context} and $\tau_i$ is the observation date. Different embeddings may come from different foundation models depending on the demand. Multiple embeddings may also correspond to the same date.
The query date $\tau^{*}$ specifies when to forecast, and we can also choose which target
foundation model that the future embedding $\mathbf{e}^{*}\in\mathbb{R}^{d}$ lies
in.

The task is to
model the conditional distribution of the future embedding $\mathbf{e}^{*}$:
\begin{equation}p_{\theta}\bigl(\mathbf{e}^{*}\mid\mathcal{C},\tau^{*}\bigr).
  \label{eq:objective}
\end{equation}

\noindent In practice, many unobserved factors, such as human activity, can lead to different future states
for the same observation context and query date. We therefore model it as an ensemble
of multiple embeddings rather than a single regression
output. Specifically, we introduce a Gaussian latent variable
$\boldsymbol{\epsilon}$ and a conditional generator $G_{\theta}$, which maps
each latent draw to a possible future embedding given $\mathcal{C}$ and
$\tau^{*}$:
\begin{equation}
  \hat{\mathbf{e}}^{*} =
  G_{\theta}(\boldsymbol{\epsilon};\mathcal{C},\tau^{*}),
  \qquad
  \boldsymbol{\epsilon}\sim\mathcal{N}(\mathbf{0},\mathbf{I}_d).
  \label{eq:generative}
\end{equation}
where resampling
$\boldsymbol{\epsilon}$ produces multiple future embeddings. 

However, sampling alone cannot answer the question of which ensemble member is the one we are most confident of, and how much uncertainty exists in each member of the ensemble. Existing work tackles this problem by postprocessing, say, 100 samples \cite{ding2026simdiff,li2024transformer}, which is computationally ineffecient. 
To increase efficiency, we pair $G_{\theta}$ with a parametric surrogate
\begin{equation}
  q_{\theta}\bigl(\mathbf{e}^{*}\mid\mathcal{C},\tau^{*}\bigr)
  =\mathcal{N}\bigl(\boldsymbol{\mu}_{\theta},
  \mathrm{diag}(\boldsymbol{\sigma}_{\theta}^{2})\bigr),
  \label{eq:surrogate}
\end{equation}
whose mean is the point forecast and diagonal covariance provides the uncertainty information directly (\cref{sec:inference}). Together, the model $f_{\theta}$ reads a token sequence $\mathbf{T}$ built from the observation context, the query
date and a noised copy $\mathbf{z}_t$ of the future embedding at flow time $t$ \cite{li2025jit},
and in a single forward outputs the Gaussian parameters and an estimate of the clean future embedding:
\begin{equation}
  f_{\theta}\bigl(\mathbf{T}(\mathcal{C},\tau^{*};\mathbf{z}_t,t)\bigr)
  = \bigl(\boldsymbol{\mu}_{\theta},\boldsymbol{\sigma}_{\theta},\hat{\mathbf{z}}_1\bigr),
  \label{eq:unified}
\end{equation}
where $\boldsymbol{\mu}_{\theta},\boldsymbol{\sigma}_{\theta}$ depend only on $(\mathcal{C},\tau^{*})$ and are computed once per forecast, while $\hat{\mathbf{z}}_1$ estimates $\mathbf{z}_1=\mathbf{e}^{*}$ from $(\mathbf{z}_t,t)$ and is re-evaluated at every solver step to generate samples $\hat{\mathbf{e}}^{*}$ from $\boldsymbol{\epsilon}$ (\cref{sec:inference}).
\subsection{Token space design}
\label{sec:backbone}
We convert all the elements of $(\mathcal{C},\tau^{*};\mathbf{z}_t,t)\bigr)$
into a sequence of $D$-dimensional tokens for batch transformer inputs. $D$ is decided by the choice of embedding space.
\begin{itemize}
  \item \textbf{Location token.}
    Seasonal changes and periodic activities vary across regions, and
    location information enables the model to learn these region-specific
    temporal patterns. We encode the imaging location with the random Fourier features (RFF) of GeoCLIP~\cite{cepeda2023geoclip}, chosen after comparing against other location encoders (\cref{sec:exp}). RFF has no trainable weights and is computed as follows:

    \begin{equation}
    \mathbf{u}(\ell)=
    \bigl[\,\cos(\boldsymbol{\Omega}^{\top}\mathbf{x}),\,
    \sin(\boldsymbol{\Omega}^{\top}\mathbf{x})\,\bigr]
    \in\mathbb{R}^{D},
    \label{eq:rff}
    \end{equation}
    where $\mathbf{x}=(\lambda,\varphi)$ is the location in radians and
    $\boldsymbol{\Omega}\in\mathbb{R}^{2\times D/2}$ is a frequency matrix
    with entries $\Omega_{jk}\sim\mathcal{N}(0,\sigma^{2})$.
    The output in \cref{eq:rff} is directly usable as a single
    token without any projection.

  \item \textbf{Date encoding and query date token.}
    Observation dates provide essential temporal context for modelling how embeddings evolve. Here we represent dates using relative observation order for
    temporal order and day-of-year (DOY) for seasonality \cite{feng2026tessera}. With the $N$ historical observations sorted by date,
    the $i$-th one has $r=i-N-1\in\{-N,\dots,-1\}$ and the forecast date has $r=0$, which expresses its
    order relative to the forecast date, avoiding dependence on absolute calendar
    years. We use sine and cosine features to represent both the order
    across observations and variation within a year \cite{herzog2025olmoearth}, allocating half of
    the $D$ dimensions to each. Each component contains $B=D/4$
sine--cosine pairs. Their scales are
    $\omega_j=10^{4j/(B-1)}$, $j=0,\dots,B-1$, spaced geometrically from
    $1$ to $10^{4}$ to cover both faster and slower temporal variation.
    The two components are
    \begin{align}
    \boldsymbol{\psi}^{\mathrm{ord}}(\tau) &=
    \Bigl[\sin\tfrac{r}{\omega_j}\;\Vert\;\cos\tfrac{r}{\omega_j}\Bigr]_{j},
    \label{eq:year}\\
    \boldsymbol{\psi}^{\mathrm{doy}}(\tau) &=
    \Bigl[\sin\bigl(\eta\sqrt{\omega_j}\bigr)\;\Vert\;
    \cos\bigl(\eta\sqrt{\omega_j}\bigr)\Bigr]_{j},
    \label{eq:doy}
    \end{align}
    where $\eta=2\pi(\mathrm{DOY}(\tau)-1)/365.25$ is the seasonal phase
    and $\Vert$ denotes concatenation. In \cref{eq:year}, dividing by
    $\omega_j$ combines sensitivity to differences between adjacent observations with
    more slowly varying features for longer-term trends. In
    \cref{eq:doy}, multiplying the seasonal phase by $\sqrt{\omega_j}$
    instead provides frequencies from $1$ to $100$. The lowest frequency
    represents the annual cycle, while higher frequencies describe finer
    within-year variation. Concatenating
    \cref{eq:year,eq:doy} gives $\boldsymbol{\psi}(\tau)\in\mathbb{R}^{D}$.
    The query date $\tau^{*}$ is encoded in the same way and passed through its own learned projection to give the query date token $\boldsymbol{\psi}^{*}$.

  \item \textbf{Historical observation tokens.}
    Although \method{} uses ready-to-use Earth embeddings, they need to be temporally ordered. Hence, we append a date encoding to each historical embedding before it is used as an input to the model. Two separate learned projections subsequently convert this input into a shared feature space, as described next. Let $\mathbf{W}_{s}\in\mathbb{R}^{D\times d_s}$ map 
    embeddings from foundation model $s$ and let $\mathbf{W}_{\tau}\in\mathbb{R}^{D\times D}$ project the date encoding. Then, the resulting observation token is
    \begin{equation}
    \mathbf{x}_i = \mathbf{W}_{s_i}\,\mathbf{e}_i
      + \mathbf{W}_{\tau}\,\boldsymbol{\psi}(\tau_i),
    \label{eq:obstoken}
    \end{equation}

  \item \textbf{Gaussian mean/std tokens.} Since a generated future embedding is just one plausible future, we append two learned tokens to the transformer input: the \emph{Gaussian mean} token $\mathbf{h}_{\mu}$ and the \emph{Gaussian std}
    token $\mathbf{h}_{\sigma}$. Their transformer outputs
    $\mathbf{o}_{\mu},\mathbf{o}_{\sigma}$ are decoded by two linear heads into
    the parameters of the surrogate \cref{eq:surrogate},
    \begin{equation}
    \boldsymbol{\mu}=\mathbf{W}_{\mu}\mathbf{o}_{\mu},\qquad
    \boldsymbol{\sigma}=\varsigma(\mathbf{W}_{\sigma}\mathbf{o}_{\sigma})+\sigma_{\min},
    \label{eq:gaussheads}
    \end{equation}
    where $\mathbf{W}_{\mu} \mathbf{W}_{\sigma}\in\mathbb{R}^{d\times D}$ are learnable linear heads and $\sigma_{\min}>0$ is a floor to keep the log-density of \cref{eq:nll}
    bounded. These tokens, together
    with the location, history and query-date tokens, form the
    \emph{probability-estimation branch} of the sequence (see \ref{fig:overview}).

  \item \textbf{Embedding-generation tokens.}
    Three further tokens serve the embedding-generation branch. A \emph{noisy target}
    token $\mathbf{h}_{\mathrm{noise}}$ carries a noised copy $\mathbf{z}_t$
    of the future embedding, and a \emph{flow time} token
    $\mathbf{h}_{\mathrm{time}}$ carries its noise level $t$. To prevent the
    model from stealing from historical observations when forecasting, we
    append a learned \emph{query} token
    $\mathbf{h}_{\mathrm{query}}\in\mathbb{R}^{D}$ and use its output as the
    forecast readout. It is initialised with small Gaussian noise, and its readout is mapped to the $d$-dimensional target
    embedding space. These
    three tokens form the \emph{embedding-generation branch} of the sequence (see \ref{fig:overview}).
\end{itemize}

\subsection{Joint training of probability estimation and embedding generation}
\label{sec:flow}
\method{} simultaneously trains both branches of the network. The
\emph{probability-estimation} branch directly predicts the mean and standard
deviation of the future embedding distribution. The \emph{embedding-generation} branch
learns to draw plausible futures by flow matching. Both are read out from a
single masked pass of \cref{eq:unified}.

\noindent\textbf{Probability-estimation branch.}
The Gaussian tokens predict $\boldsymbol{\mu}$ and $\boldsymbol{\sigma}$ via
\cref{eq:gaussheads}. We train them by minimising the negative log-likelihood
of the future embedding under the surrogate \cref{eq:surrogate}:
\begin{equation}
  \mathcal{L}_{\mathrm{nll}}=\frac{1}{d}\sum_{k=1}^{d}\Bigl[
    \frac{(e^{*}_k-\mu_k)^{2}}{2\sigma_k^{2}}
    +\log\sigma_k\Bigr]+\frac{1}{2}\log 2\pi .
  \label{eq:nll}
\end{equation}
The first term scales prediction errors by their estimated uncertainty, while
$\log\sigma_k$ prevents unbounded variance. This branch therefore provides a
point forecast and uncertainty directly, without sampling or integration. Note that Gaussian NLL does not require the true conditional distribution to be Gaussian when learning the moments.

\noindent\textbf{Embedding-generation branch.}
Rectified flow~\cite{liu2023rectified,lipman2023flowmatching} has a
straight path from Gaussian noise $\boldsymbol{\epsilon}$ to the target
embedding $\mathbf{z}_1$. At a randomly sampled flow time $t$, we construct
\begin{equation}
  \mathbf{z}_t = t\,\mathbf{z}_1 + (1-t)\,\boldsymbol{\epsilon},
  \qquad
  \boldsymbol{\epsilon}\sim\mathcal{N}(\mathbf{0},\mathbf{I}_d),
  \quad t\in[0,1].
  \label{eq:interp}
\end{equation}
The model receives $\mathbf{z}_t$ and $t$ through the noisy-target and
flow-time tokens
\begin{equation}
  \mathbf{h}_{\mathrm{noise}} = \mathbf{W}_{z}\,\mathbf{z}_t,
  \qquad
  \mathbf{h}_{\mathrm{time}} = \mathrm{MLP}\bigl(\boldsymbol{\zeta}(t)\bigr),
  \label{eq:noisytoken}
\end{equation}
where $\boldsymbol{\zeta}(t)$ is a sinusoidal encoding. Since Earth embeddings
are high-dimensional signals, the query-token output acts as a denoiser and
predicts the clean future embedding rather than the velocity~\cite{li2025jit};
the path velocity $\mathbf{v}_{\theta}$ then follows,
\begin{equation}
  \hat{\mathbf{z}}_1=\mathbf{W}_{\mathrm{query}}\,\mathbf{o}_{\mathrm{query}},
  \qquad
  \mathbf{v}_{\theta}(\mathbf{z}_t,t;\mathcal{C},\tau^{*})
  =\frac{\hat{\mathbf{z}}_1-\mathbf{z}_t}{1-t},
  \label{eq:velocity}
\end{equation}
where $1-t$ is clipped at a small constant.
We train the branch by matching the predicted velocity to the straight-path
target $\mathbf{z}_1-\boldsymbol{\epsilon}$:
\begin{equation}
  \mathcal{L}_{\mathrm{flow}}=\frac{1}{d}\,\mathbb{E}_{t,\boldsymbol{\epsilon}}
  \bigl\lVert \mathbf{v}_{\theta}(\mathbf{z}_t,t;\mathcal{C},\tau^{*})
              -(\mathbf{z}_1-\boldsymbol{\epsilon}) \bigr\rVert_2^{2}.
  \label{eq:loss}
\end{equation}

\noindent\textbf{Mixed objective.}
We jointly optimise the probability-estimation and embedding-generation
branches as
\begin{equation}
  \mathcal{L}=\mathcal{L}_{\mathrm{nll}}
  + \mathcal{L}_{\mathrm{flow}},
  \label{eq:total}
\end{equation}
Both branches are evaluated in one transformer
pass, using an asymmetric mask that prevents probability-estimation tokens
from attending to embedding-generation tokens. Both loss terms have the same scale, and equal weight.
\subsection{Inference and anomaly change detection}
\label{sec:inference}
The trained network supports forecasting and forecast-based anomaly change detection without retraining or task-specific heads.

\noindent\textbf{Forecasting.}
Given any query date $\tau^{*}>\tau_N$, \method{} accepts the available
context $\mathcal{C}$ and returns a point forecast, $J$ ensemble forecasts,
and their predictive uncertainty as one tuple:
\begin{equation}
  \begin{aligned}
  \mathcal{F}_{\theta}(\mathcal{C},\tau^{*})
    &=\left(\hat{\mathbf{e}}^{*}_{\mathrm{pt}},
      \{\hat{\mathbf{e}}^{*(c)}\}_{c=1}^{J},
      \boldsymbol{\sigma}\right),\\
  \hat{\mathbf{e}}^{*}_{\mathrm{pt}}
    &=\frac{\boldsymbol{\mu}}{\lVert\boldsymbol{\mu}\rVert_2},
  \qquad
  \hat{\mathbf{e}}^{*(c)}
    =\frac{\mathbf{y}^{(c)}_1}{\lVert\mathbf{y}^{(c)}_1\rVert_2}.
  \end{aligned}
  \label{eq:forecast}
\end{equation}
The Gaussian branch obtains $(\boldsymbol{\mu},\boldsymbol{\sigma})$ in one
pass with $\mathbf{z}_t=\mathbf{0}$ and $t=0$. For each ensemble forecast,
$\mathbf{y}^{(c)}_0=\boldsymbol{\epsilon}^{(c)}\sim
\mathcal{N}(\mathbf{0},\mathbf{I}_d)$ is advanced using $S$ Euler steps,
\begin{equation}
  \mathbf{y}^{(c)}_{(n+1)/S}=\mathbf{y}^{(c)}_{n/S}
  +\tfrac{1}{S}\mathbf{v}_{\theta}\!\left(
  \mathbf{y}^{(c)}_{n/S},\tfrac{n}{S};\mathcal{C},\tau^{*}\right),
  \quad n=0,\ldots,S-1,
  \label{eq:euler}
\end{equation}
and all $J$ paths are processed in one batch. Thus, the same model can query
arbitrary future dates, provide a fast point forecast with per-dimension
uncertainty, and represent multiple possible futures when needed.

\noindent\textbf{Anomaly change detection.}
After observing $\mathbf{e}^{*}_{\mathrm{obs}}$ at $\tau^{*}$, we compare it
with both the Gaussian forecast and the closest ensemble forecast, scaled by the
predicted uncertainty:
\begin{equation}
  a = \min\Bigl\{ \Bigl\lVert
      \frac{\mathbf{e}^{*}_{\mathrm{obs}}-\boldsymbol{\mu}}{\boldsymbol{\sigma}}
      \Bigr\rVert_2 ,\;
    \min_{c\in\{1,\ldots,J\}} \Bigl\lVert
      \frac{\mathbf{e}^{*}_{\mathrm{obs}}-\hat{\mathbf{e}}^{*(c)}}{\boldsymbol{\sigma}}
      \Bigr\rVert_2 \Bigr\},
  \label{eq:anomaly}
\end{equation}
where division is element-wise. Taking the minimum, the score is high only
when the observation is far from both the mean prediction and all ensemble
forecasts, \ie a change that any plausible forecast explains is treated as
normal, while
$\boldsymbol{\sigma}$ discounts expected variation. We use $J\!=\!10$ and
$S\!=\!10$ at default; applying \cref{eq:anomaly} across locations yields an anomaly map
without change labels or a separately trained detection head.

\section{Results}
\label{sec:exp}

We evaluate \method{} on two tasks: forecasting future embeddings from irregular histories, and detecting anomalous changes by comparing new observations with these forecasts.
We first describe the datasets, baselines, metrics, and implementation details (\cref{sec:exp_setup,sec:exp_baselines,sec:impl}).
We then compare embedding-space and image-space forecasting and report anomaly change detection results (\cref{sec:exp_main}).
Finally, \cref{sec:exp_analysis} reports related sensitivity and ablation results.

\subsection{Datasets}
\label{sec:exp_setup}

\begin{figure*}[t]
  \centering
  \includegraphics[width=0.95\textwidth]{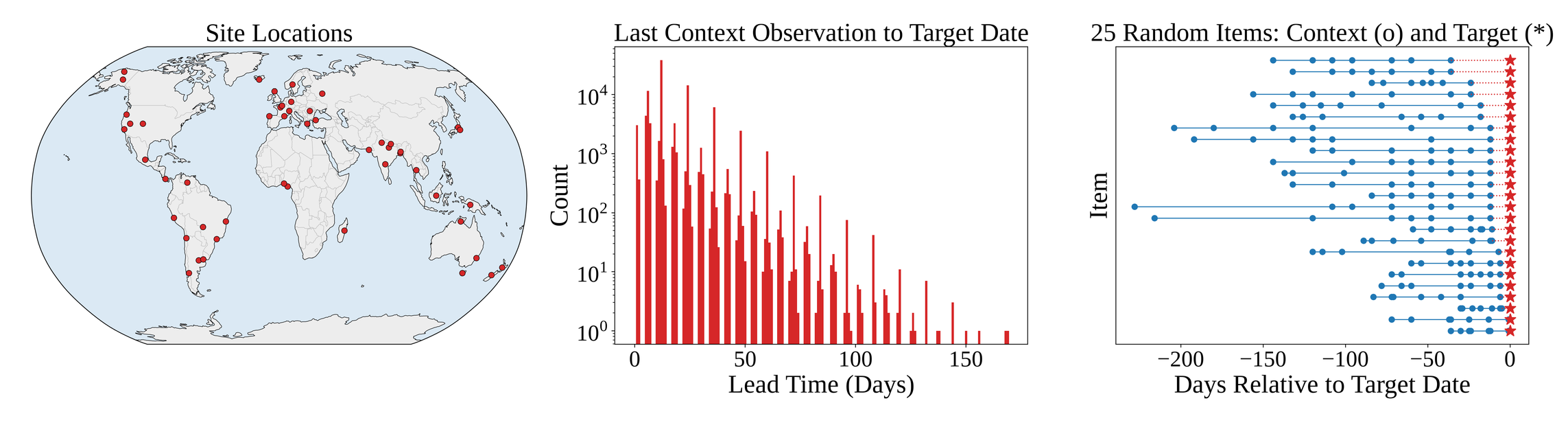}\\[1pt]
  {\small (a) Statistics on S1\_Embed}\\[3pt]
  \includegraphics[width=0.95\textwidth]{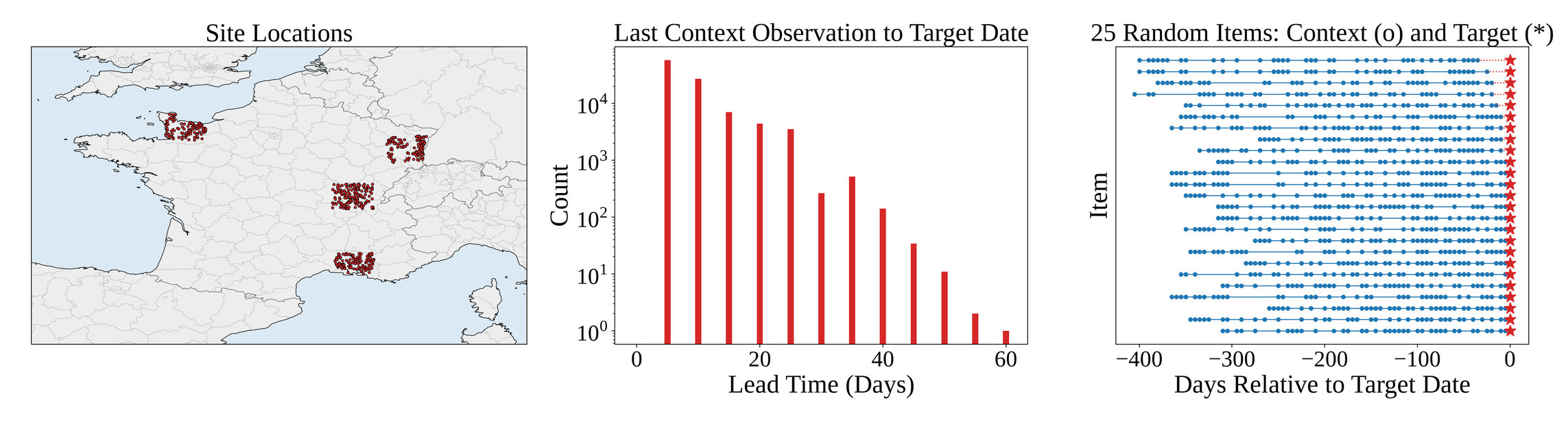}\\[1pt]
  {\small (b) Statistics on S2\_Embed}\\[3pt]
  \includegraphics[width=0.95\textwidth]{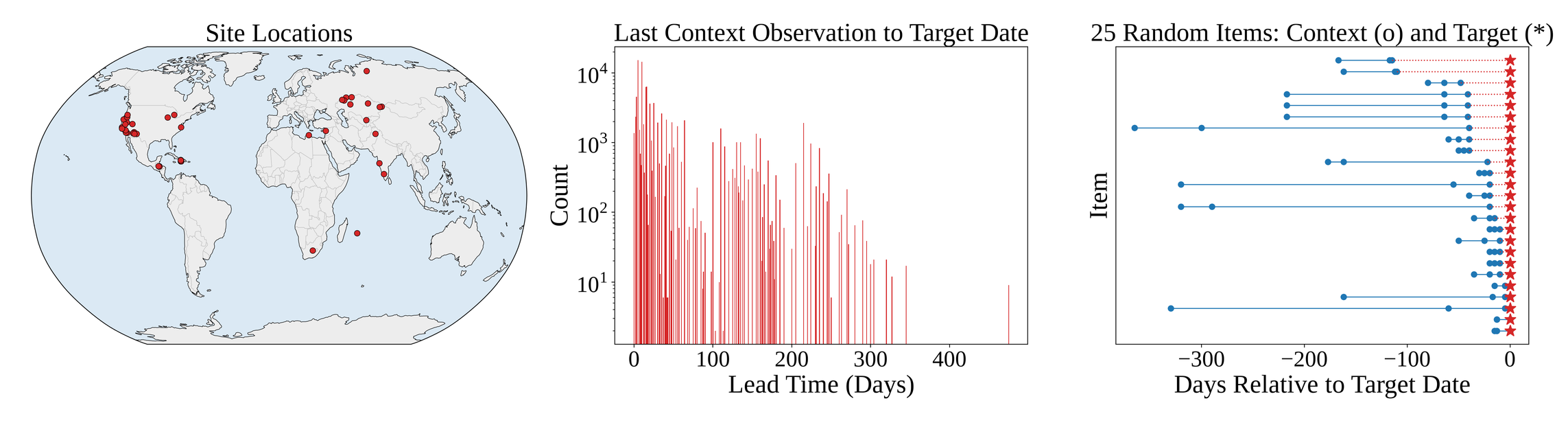}\\[1pt]
  {\small (c) Statistics on GSA}
  \caption{\textbf{Site locations and irregular temporal sampling in the S1\_Embed,
  S2\_Embed and GSA test sets.} Left: site locations. Middle: distribution of
  the lead time from the last context observation to the target date. Right:
  context dates (circles) and target date (star) of 25 random samples, relative
  to the target date. Context intervals and lead times vary across samples,
  spanning 1--170 days for S1\_Embed, 5--60 days for S2\_Embed and over a year
  for GSA, so models must forecast embeddings for an arbitrary target date.}
  \label{fig:intervals}
\end{figure*}

\paragraph{S1\_Embed for forecasting.}

This dataset forecasts embeddings of Sentinel-1 SAR time series under a
cross-year condition. It contains 60 sites of $10\times10$\,km distributed over
the globe (\cref{fig:intervals}a), 43 covering diverse biomes and 17 in large cities. Every descending
Sentinel-1 RTC acquisition of these sites in 2024 and 2025 is converted into
OlmoEarth-nano embeddings~\cite{herzog2025olmoearth} with patch size 5, \ie a
$200\times200$ grid of 128-dimensional vectors at 50\,m. Models are trained on
the 2024 series of all sites (1{,}991 scenes) and evaluated on 2025, with 10
sites (7 biome, 3 urban; 522 scenes) for validation and the remaining 50 sites
(2{,}426 scenes) for testing. A total of 1M time-series samples, 20K per test
site, are drawn to compute the metrics. Each sample forecasts with 7 earlier observations, which was chosen by the sensitivity analysis in \cref{tab:history}. Notice that both the context intervals and the lead time to the target date are irregular (\cref{fig:intervals}a).

\paragraph{S2\_Embed for forecasting.}
This dataset forecasts embeddings of optical multispectral time series under a
cross-region condition. We use PASTIS~\cite{garnot2021panoptic},
which contains 2{,}433 sites of $128\times128$ pixels in four regions of France (\cref{fig:intervals}b), each
observed by Sentinel-2 L2A on 38--61 dates spanning 2018--2019. Every date is
encoded independently by Prithvi-EO-2.0-300M~\cite{szwarcman2024prithvi},
yielding an $8\times8$ grid of 1024-dimensional tokens that each cover
$16\times16$ pixels (160\,m). Sites are split 6:2:2 into 1{,}461 training
(69{,}066 scenes), 486 validation (23{,}273 scenes) and 486 test patches
(22{,}883 scenes). A total of 1ooK time-series samples are drawn from the test
patches to compute the metrics. A sample forecasts one token at a random target date from 37 randomly drawn earlier
dates, again not necessarily consecutive, which yields irregular context
intervals and lead times as well (\cref{fig:intervals}b).

We do not use the classical EarthNet2021~\cite{requenamesa2021earthnet}, as it
provides only RGB and near-infrared bands, while existing foundation models need complete channels to generate dense embeddings.

\paragraph{GSA for anomaly change detection.}
GSA~\cite{li2026esia} is a global dataset for anomaly change detection. It covers
19{,}801.60\,km$^2$ and 55 events from six categories: dam break, burned area,
fire point, marine debris, oil leak and water bloom. For each event, one
post-event Sentinel-2 L2A image and a varying number of pre-event images form a monitoring stream, whose sites and irregular intervals are shown in \cref{fig:intervals}c, and the post-event image carries a
pixel-level binary anomaly annotation. Anomalies are rare and small: most events
have an anomaly ratio below $10\%$. We have tested \method{} with different embedding spaces on this dataset, and additionally collect annual Tessera 1.1 ~\cite{feng2026tessera} and
AlphaEarth~\cite{brown2025alphaearth} embeddings in the past three years as extra
conditions.

\begin{table*}[!t]
  \centering
  \small
  \setlength{\tabcolsep}{4pt}
  \newcommand{\hd}[1]{\begin{tabular}[c]{@{}c@{}}#1\end{tabular}}
  \newcommand{\cat}[1]{\multicolumn{11}{@{}l}{\textit{\color{gray}#1}}}
  \begin{tabular*}{\linewidth}{@{\extracolsep{\fill}}l|ccccc|ccccc@{}}
    \toprule
    \multirow{2}{*}{Method}
      & \multicolumn{5}{c|}{S1\_Embed} & \multicolumn{5}{c}{S2\_Embed} \\
      & \hd{Mean $z$\\$\rightarrow0$} & \hd{Std $z$\\$\rightarrow1$} & \hd{Point\\Cosine $\downarrow$} & \hd{Ensemble\\Cosine $\downarrow$} & CRPS $\downarrow$
      & \hd{Mean $z$\\$\rightarrow0$} & \hd{Std $z$\\$\rightarrow1$} & \hd{Point\\Cosine $\downarrow$} & \hd{Ensemble\\Cosine $\downarrow$} & CRPS $\downarrow$ \\
    \midrule
    \cat{Deterministic Models} \\
    Persistence
      & -- & -- & 0.1760 & 0.1760 & --
      & -- & -- & 0.1699 & 0.1699 & -- \\
    ETSformer~\cite{woo2022etsformer}
      & -- & -- & 0.1109 & 0.1109 & --
      & -- & -- & 0.0737 & \underline{0.0737} & -- \\
    Earthformer~\cite{gao2022earthformer}
      & -- & -- & 0.1103 & 0.1103 & --
      & -- & -- & 0.0808 & 0.0808 & -- \\
    \midrule
    \cat{Image-Space Forecast with Designed Query-Date} \\
    DeepAR~\cite{salinas2020deepar}
      & $-0.0022$ & 3.2313 & 0.1706 & \underline{0.0926} & 0.0281
      & $-0.0184$ & 1.7303 & 0.1812 & 0.2287 & 0.0073 \\
    TFT~\cite{lim2021temporal}
      & 0.0018 & 4.0660 & 0.1698 & 0.1039 & 0.0311
      & $-0.0189$ & 2.3262 & 0.2359 & 0.2539 & 0.0086 \\
    SimDiff~\cite{ding2026simdiff}
      & $-0.0164$ & 2.3214 & 0.2061 & 0.1491 & 0.0359
      & $-0.0012$ & 2.0941 & 0.2454 & 0.1757 & 0.0072 \\
    \midrule
    \cat{Embedding-Space Forecast with Designed Token Space} \\
    DeepAR$^*$~\cite{salinas2020deepar}
      & $\mathbf{-0.0001}$ & \textbf{1.0771} & \underline{0.1089} & 0.1439 & 0.0220
      & $\mathbf{0.0005}$ & \underline{1.0341} & \underline{0.0679} & 0.0755 & \underline{0.0045} \\
    TFT$^*$~\cite{lim2021temporal}
      & $-0.0449$ & 1.1776 & 0.1101 & 0.1367 & \underline{0.0217}
      & $-0.0424$ & 1.0924 & 0.0951 & 0.1097 & 0.0052 \\
    TMDM$^*$~\cite{li2024transformer}
      & 0.0062 & 0.6874 & 0.1210 & 0.3211 & 0.0308
      & 0.0072 & 0.4925 & 0.0948 & 0.3359 & 0.0082 \\
    SimDiff$^*$~\cite{ding2026simdiff}
      & $-0.0046$ & 1.4352 & 0.1105 & 0.1019 & 0.0236
      & 0.0020 & 1.4991 & 0.0947 & 0.0876 & 0.0051 \\
    \method{} (Ours)
      & \underline{$-0.0005$} & \underline{1.1376} & \textbf{0.1081} & \textbf{0.0911} & \textbf{0.0213}
      & \underline{0.0006} & \textbf{1.0250} & \textbf{0.0667} & \textbf{0.0613} & \textbf{0.0038} \\
    \bottomrule
  \end{tabular*}
  \caption{\textbf{Forecasting results on S1\_Embed and S2\_Embed.} Metrics are
  defined in \cref{sec:exp_setup}. Deterministic models give a single forecast,
  so only cosine distances apply. $^*$ marks baselines substantially adapted to
  the embedding space. \textbf{Bold}: best; \underline{underlined}: second best.}
  \label{tab:main}
\end{table*}

\begin{figure*}[t]
  \centering
  \includegraphics[width=\linewidth]{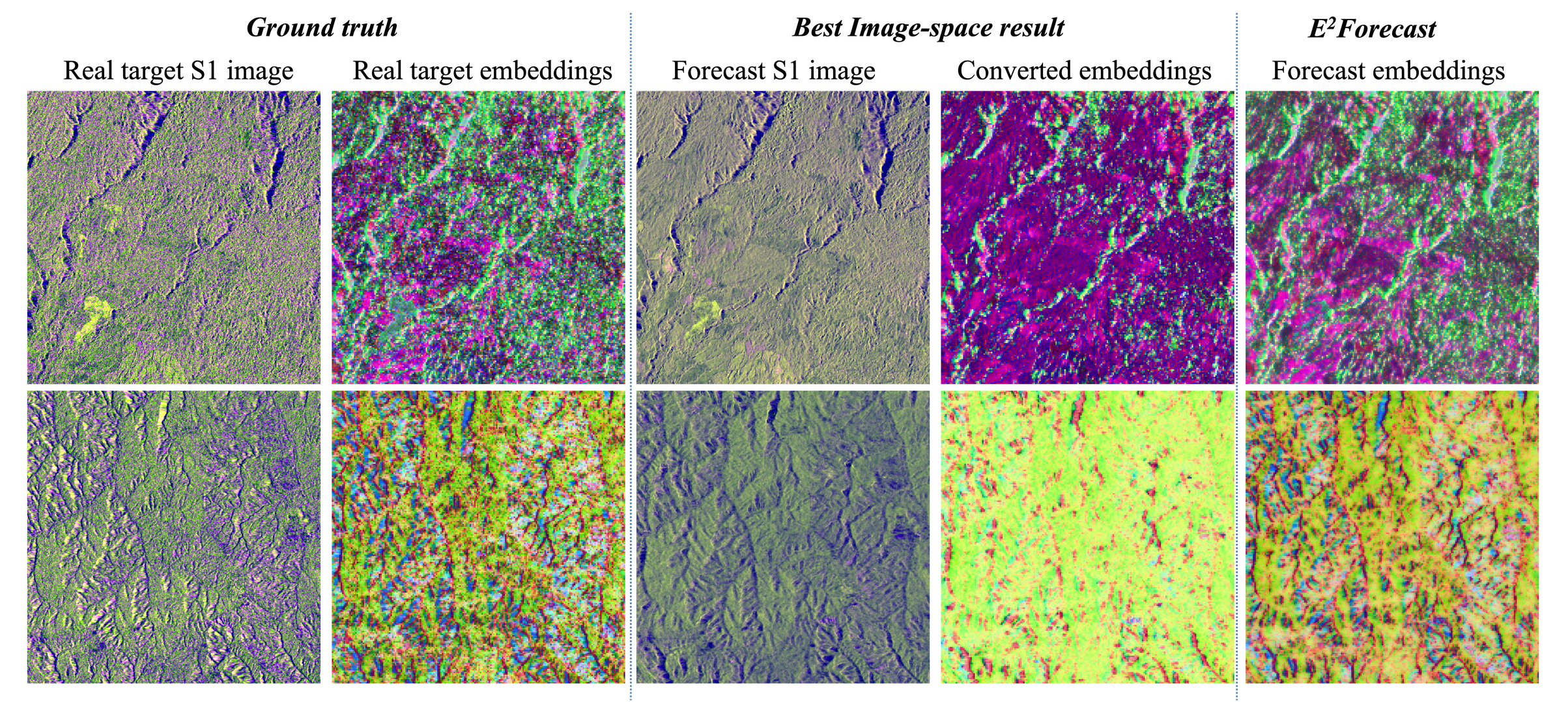}
  \caption{\textbf{Image-space \vs embedding-space forecasting on S1\_Embed.}
  For each example, we show the target image with R=VV, G=VH, and B=VV-VH. Embeddings are visualized with the first three PCA components.}
  \label{fig:image_vs_embedding_s1}
\end{figure*}

\begin{figure*}[t]
  \centering
  \includegraphics[width=\linewidth]{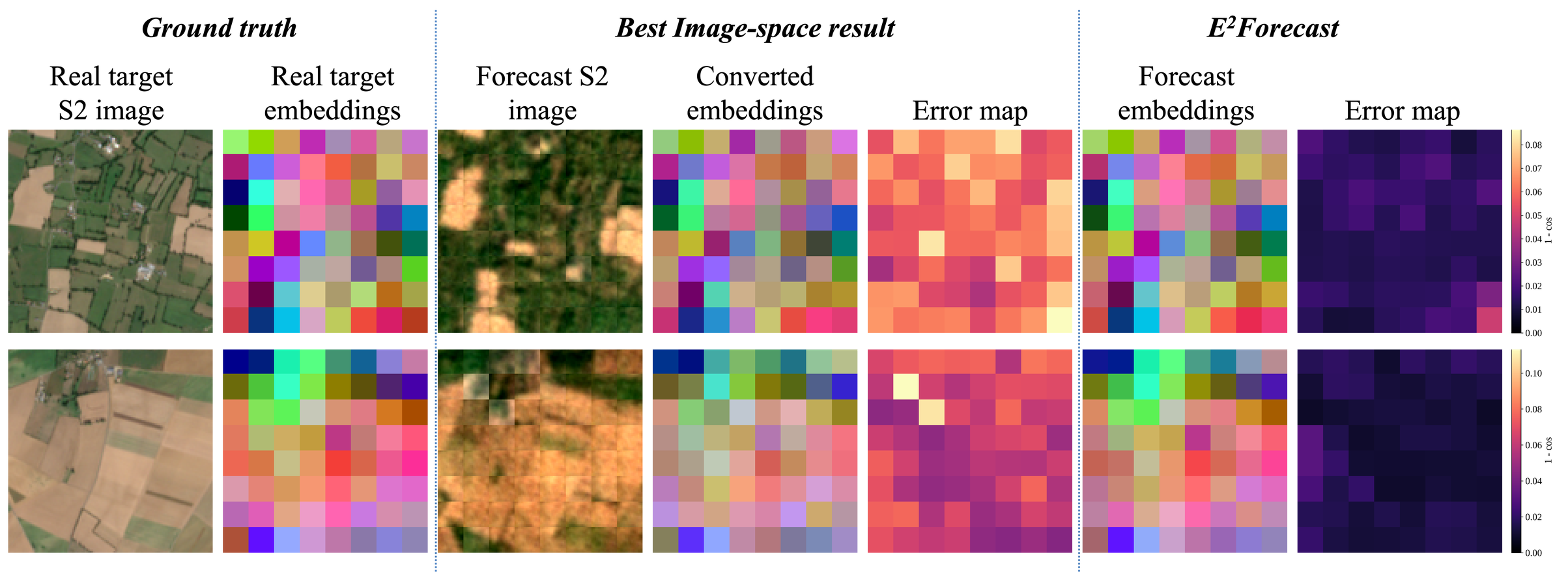}
  \caption{\textbf{Image-space \vs embedding-space forecasting on S2\_Embed.} We show the RGB for the S2 image and visualize embeddings with the first three PCA components. The error maps are computed with error maps of $1-\cos$ over all embedding dimensions.}
  \label{fig:image_vs_embedding_s2}
\end{figure*}

\subsection{Baselines and metrics}
\label{sec:exp_baselines}
We compare \method{} with seven forecasting models that cover different output types: the deterministic Persistence,
which takes the closest observation as the forecast,
ETSformer~\cite{woo2022etsformer} and Earthformer~\cite{gao2022earthformer}; the probabilistic
DeepAR~\cite{salinas2020deepar} and Temporal Fusion Transformer
(TFT)~\cite{lim2021temporal}; and the generative TMDM~\cite{li2024transformer}
and SimDiff~\cite{ding2026simdiff}. Since none of these models was designed for
embedding-space forecasting at an arbitrary date, we adapt them in two ways.
In the image space, DeepAR, TFT and SimDiff forecast the image at the query
date given our designed query-date token, and the forecast image is then
converted into embeddings by the same foundation model for evaluation.
In the embedding space, DeepAR, TFT, TMDM and SimDiff share the proposed token
space with \method{}.

We evaluate forecasting performance from three aspects:
\begin{enumerate}
    \item \textit{Uncertainty} is assessed by the mean and standard
deviation of the z-score $z=(e^{*}-\mu)/\sigma$ over all pixels and dimensions,
whose ideal values are 0 and 1.
\item A \textit{point forecast} is assessed by the point
cosine distance $1-\cos(\boldsymbol{\mu},\mathbf{e}^{*})$ between the predicted
mean and the ground truth. For generative models, we take the mean of 100 samples as the point forecast.
\item \textit{Ensemble forecasts} are assessed by the ensemble
cosine distance, \ie the smallest $1-\cos(\hat{\mathbf{e}}^{*},\mathbf{e}^{*})$
among the ensemble members, and by the CRPS~\cite{gneiting2007strictly}. When computing the ensemble cosine distance, we sample 100 samples for models that only output a Gaussian (DeepAR and TFT) and generate 10 forecasts for generative models (including \method{}). Flow and diffusion models use 10 Euler and 50 solver steps to generate
one sample, respectively. Besides, we also compare with the persistence setting as a baseline, which takes the closest past observation as the forecast.
\end{enumerate}

For anomaly change detection, using the GSA dataset,
we compare with the 23 models benchmarked in~\cite{li2026esia}: 14
category-specific spectral-index detectors (\eg dNBR and FAI), the bi-temporal
change detectors CVA~\cite{bovolo2007cva}, SCG~\cite{sun2021scg},
MAD~\cite{nielsen2007irmad} and SAM-Bi~\cite{li2026anomalycd}, and the
time-series detectors COLD~\cite{zhu2020cold}, EOAD~\cite{castillo2021eoad},
PKAD~\cite{wei2023pkad}, AnomalyCD~\cite{li2026anomalycd} and
ESIA~\cite{li2026esia}. Following~\cite{li2026esia}, we report \textit{pixel-level
precision, recall, and $\mathrm{F}_1$ }of the binary anomaly maps against the annotations.

\begin{table*}[!t]
  \centering
  \footnotesize
  \setlength{\tabcolsep}{2.5pt}
  \newcommand{\cat}[1]{\multicolumn{13}{@{}l}{\textit{\color{gray}#1}}}
  \begin{tabular*}{\linewidth}{@{\extracolsep{\fill}}lccc|lccc|lccc|c@{}}
    \toprule
    \multicolumn{4}{c|}{Dam break} & \multicolumn{4}{c|}{Burned area}
      & \multicolumn{4}{c|}{Fire point} & \multirow{2}{*}{Input type} \\
    Model & R & P & F$_1$ & Model & R & P & F$_1$ & Model & R & P & F$_1$ & \\
    \midrule
    \cat{Specialized anomaly localization models} \\
    MNDWI~\cite{xu2006mndwi} & 74.64 & 37.33 & 40.37 & dNDVI~\cite{jiang2006ndvi} & 39.01 & \textbf{87.32} & 48.24 & TAI~\cite{xia2018tai} & 2.91 & 14.81 & 3.92 & Reflectance \\
    MBWI~\cite{wang2018mbwi} & 62.59 & 42.52 & 41.64 & dNBR~\cite{lutes2006firemon} & 43.30 & \underline{82.05} & 52.50 & NDFI~\cite{liu2025ndfi} & 24.64 & \underline{49.69} & 21.72 & Reflectance \\
    AWEI~\cite{jiang2014awei} & 69.21 & 43.73 & 43.44 & RdNBR~\cite{miller2007rdnbr} & 63.97 & 67.10 & 62.13 & AFD-S$^2$~\cite{hu2021afds2} & 2.33 & \textbf{89.95} & 4.51 & Reflectance \\
    \midrule
    \cat{General change detection models (bi-temporal)} \\
    CVA~\cite{bovolo2007cva} & 33.65 & 58.90 & 29.82 & CVA~\cite{bovolo2007cva} & 15.90 & 71.42 & 23.56 & CVA~\cite{bovolo2007cva} & 40.68 & 36.97 & 31.35 & Reflectance \\
    SCG~\cite{sun2021scg} & 62.83 & 43.04 & 38.97 & SCG~\cite{sun2021scg} & 38.36 & 62.01 & 42.11 & SCG~\cite{sun2021scg} & 60.21 & 33.05 & 31.87 & Reflectance \\
    MAD~\cite{nielsen2007irmad} & 48.48 & 56.05 & 35.77 & MAD~\cite{nielsen2007irmad} & 26.78 & 71.66 & 35.73 & MAD~\cite{nielsen2007irmad} & 52.16 & 34.30 & \underline{33.22} & Reflectance \\
    SAM-Bi~\cite{li2026anomalycd} & 65.94 & 46.43 & 43.76 & SAM-Bi~\cite{li2026anomalycd} & 50.04 & 56.23 & 47.18 & SAM-Bi~\cite{li2026anomalycd} & 48.13 & 22.58 & 22.62 & SAM features \\
    \midrule
    \cat{General anomaly localization models (time-series)} \\
    COLD~\cite{zhu2020cold} & \textbf{91.06} & 39.27 & 44.71 & COLD~\cite{zhu2020cold} & 58.45 & 64.39 & 47.01 & COLD~\cite{zhu2020cold} & \textbf{67.62} & 23.69 & 22.79 & Reflectance \\
    EOAD~\cite{castillo2021eoad} & 52.21 & 38.15 & 36.44 & EOAD~\cite{castillo2021eoad} & 19.98 & 52.21 & 24.54 & EOAD~\cite{castillo2021eoad} & 53.33 & 29.56 & 30.41 & Vegetation index \\
    PKAD~\cite{wei2023pkad} & 71.15 & 39.58 & 28.69 & PKAD~\cite{wei2023pkad} & \textbf{89.31} & 57.81 & \underline{66.97} & PKAD~\cite{wei2023pkad} & \underline{62.68} & 17.35 & 20.27 & Vegetation index \\
    AnomalyCD~\cite{li2026anomalycd} & 56.70 & \textbf{66.20} & 52.24 & AnomalyCD~\cite{li2026anomalycd} & 46.92 & 68.16 & 49.38 & AnomalyCD~\cite{li2026anomalycd} & 38.37 & 35.79 & 29.63 & SAM features \\
    ESIA~\cite{li2026esia} & 62.98 & 58.77 & \underline{54.32} & ESIA~\cite{li2026esia} & \underline{82.36} & 56.04 & 63.31 & ESIA~\cite{li2026esia} & 41.60 & 31.06 & 32.09 & HyperFree features \\
    \midrule
    \method{} & \underline{87.27} & \underline{60.46} & \textbf{60.91} & \method{} & 80.76 & 81.62 & \textbf{78.38} & \method{} & 48.03 & 41.24 & \textbf{38.70} & Embeddings \\
    \bottomrule
  \end{tabular*}

  \vspace{6pt}

  \begin{tabular*}{\linewidth}{@{\extracolsep{\fill}}lccc|lccc|lccc|ccc@{}}
    \toprule
    \multicolumn{4}{c|}{Marine debris} & \multicolumn{4}{c|}{Oil leak}
      & \multicolumn{4}{c|}{Water bloom} & \multicolumn{3}{c}{Average} \\
    Model & R & P & F$_1$ & Model & R & P & F$_1$ & Model & R & P & F$_1$ & R & P & F$_1$ \\
    \midrule
    \multicolumn{15}{@{}l}{\textit{\color{gray}Specialized anomaly localization models}} \\
    FDI~\cite{ali2024fdi} & 62.08 & \underline{20.93} & \textbf{19.44} & NDOI~\cite{brekke2005ndoi} & 51.29 & 4.34 & 6.16 & FAI~\cite{hu2009fai} & 57.36 & 28.76 & 34.30 & -- & -- & -- \\
    PI~\cite{themistocleous2020pi} & 22.68 & 5.89 & 6.38 & -- & -- & -- & -- & SABI~\cite{alawadi2010sabi} & 61.21 & 39.62 & 36.93 & -- & -- & -- \\
    \midrule
    \multicolumn{15}{@{}l}{\textit{\color{gray}General change detection models (bi-temporal)}} \\
    CVA~\cite{bovolo2007cva} & 46.35 & 7.81 & 6.56 & CVA~\cite{bovolo2007cva} & 8.61 & 4.65 & 6.03 & CVA~\cite{bovolo2007cva} & 12.15 & 45.09 & 19.09 & 26.22 & 37.47 & 19.40 \\
    SCG~\cite{sun2021scg} & \textbf{84.77} & 0.73 & 1.43 & SCG~\cite{sun2021scg} & 43.05 & 6.45 & 10.52 & SCG~\cite{sun2021scg} & 52.15 & 29.72 & 29.91 & 56.90 & 29.17 & 25.80 \\
    MAD~\cite{nielsen2007irmad} & 52.43 & 4.82 & 5.13 & MAD~\cite{nielsen2007irmad} & 10.16 & 4.41 & 6.10 & MAD~\cite{nielsen2007irmad} & 19.23 & 42.07 & 26.30 & 34.87 & 35.55 & 23.71 \\
    SAM-Bi~\cite{li2026anomalycd} & 18.09 & 10.42 & 11.60 & SAM-Bi~\cite{li2026anomalycd} & 53.37 & 11.45 & 18.47 & SAM-Bi~\cite{li2026anomalycd} & 41.25 & 18.20 & 23.10 & 46.14 & 27.55 & 27.79 \\
    \midrule
    \multicolumn{15}{@{}l}{\textit{\color{gray}General anomaly localization models (time-series)}} \\
    COLD~\cite{zhu2020cold} & \underline{70.34} & 4.55 & 4.48 & COLD~\cite{zhu2020cold} & \textbf{69.64} & 3.90 & 7.21 & COLD~\cite{zhu2020cold} & 51.79 & 33.61 & 33.91 & \underline{68.15} & 28.24 & 26.69 \\
    EOAD~\cite{castillo2021eoad} & 42.85 & 0.40 & 0.78 & EOAD~\cite{castillo2021eoad} & 32.62 & 10.38 & 12.91 & EOAD~\cite{castillo2021eoad} & 38.90 & 31.49 & 28.14 & 39.98 & 27.03 & 22.20 \\
    PKAD~\cite{wei2023pkad} & 67.29 & 0.99 & 1.83 & PKAD~\cite{wei2023pkad} & \underline{66.18} & 4.23 & 7.80 & PKAD~\cite{wei2023pkad} & \textbf{83.95} & 19.80 & 28.57 & \textbf{73.43} & 23.29 & 25.69 \\
    AnomalyCD~\cite{li2026anomalycd} & 28.07 & 10.09 & 14.36 & AnomalyCD~\cite{li2026anomalycd} & 44.01 & 7.63 & 12.87 & AnomalyCD~\cite{li2026anomalycd} & 55.50 & 16.49 & 22.72 & 44.93 & 34.06 & 30.27 \\
    ESIA~\cite{li2026esia} & 34.56 & \textbf{21.26} & \underline{18.33} & ESIA~\cite{li2026esia} & 38.73 & \underline{18.11} & \underline{20.03} & ESIA~\cite{li2026esia} & 69.80 & \underline{55.30} & \underline{61.53} & 54.99 & \underline{40.05} & \underline{41.58} \\
    \midrule
    \method{} & 56.55 & 12.04 & 14.61 & \method{} & 42.87 & \textbf{23.27} & \textbf{26.01} & \method{} & \underline{77.38} & \textbf{74.39} & \textbf{73.58} & 65.48 & \textbf{48.84} & \textbf{48.70} \\
    \bottomrule
  \end{tabular*}
  \caption{\textbf{Anomaly change detection results on GSA.} Pixel-level recall
  (R), precision (P) and F$_1$ (\%) for each of the six event categories and
  their average. Baseline results are taken from~\cite{li2026esia}. Best and
  second best per column are in \textbf{bold} and \underline{underlined}.}
  \label{tab:gsa}
\end{table*}

\begin{figure}[t]
  \centering
  \includegraphics[width=\linewidth]{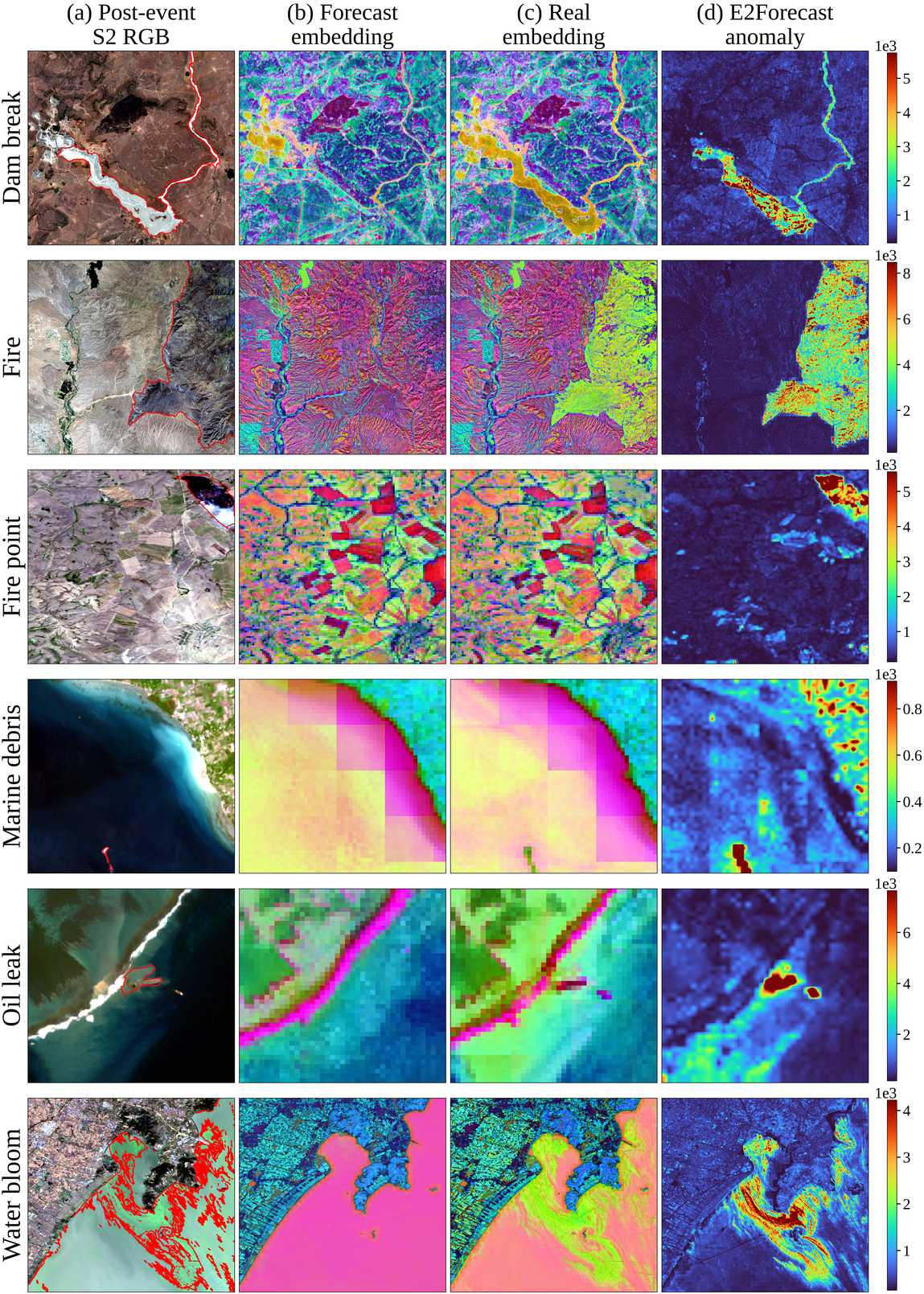}
  \caption{\textbf{Qualitative anomaly change detection results on GSA.} One
  event per category. (a) Post-event Sentinel-2 RGB image with the annotated
  anomaly outlined in red. (b) Embedding forecast by \method{} and (c)
  embedding observed on the post-event date, both projected to RGB. (d) Anomaly
  score of \cref{eq:anomaly}.}
  \label{fig:gsa_qualitative}
\end{figure}

\subsection{Implementation details}
\label{sec:impl}
The Transformer width equals the embedding dimension ($128$ for S1\_Embed and GSA, $1024$ for S2\_Embed, $256$ for the INR), so the location token enters without projection.
Every input projection, including the separate projection of the query date, is a LayerNorm followed by a linear layer.
The network has $8$ pre-norm layers with $4$ heads ($8$ for S2\_Embed), an FFN width of four times the embedding dimension, GELU activations and dropout $0.05$; the standard-deviation floor $\sigma_{\min}$ is $10^{-4}$.
The flow branch operates on the target embedding scaled by the square root of its dimension, so each dimension has unit variance, and generated samples are scaled back and $\ell_2$-normalised.
Flow times are drawn from a logit-normal distribution with mean $-1.2$ and standard deviation $1.2$, and $1-t$ is clipped at $0.05$.
We train with AdamW at a constant learning rate of $1.1\times10^{-4}$, no weight decay and gradient-norm clipping at $0.5$.
Forecasting models use $1000$ steps per epoch: $600$ epochs with batch size $256$ on S1\_Embed, $100$ epochs with batch size $64$ on S2\_Embed, and $300$ epochs with batch size $256$ on GSA.
The INR is trained for up to $300$ epochs of $2\times10^{5}$ samples with batch size $256$.
All results use $J=S=10$.

\subsection{Main results}
\label{sec:exp_main}

\paragraph{Forecasting in the image space loses much of the semantic features.}
Since no prior model supports forecasting from irregular observations at an arbitrary lead time, we adapt existing baselines by equipping them with a query token for image forecasting, and with the proposed full token space for embedding forecasting. When evaluating image-space forecasting, we first use these models to forecast images and then convert the forecasts into embeddings.
As shown in
\cref{tab:main}, most embedding-space models outperform their image-space
counterparts. The gap is especially large in std $z$ and point forecast. The image-space
baselines are strongly overconfident, with std $z$ between 1.73 and 4.07,
whereas all embedding-space models stay closer to the ideal value of 1.0. The point forecast cosine distances are mostly over 0.17, much worse than the embedding ones, around 0.10.

To illustrate this difference intuitively, we visualize two samples in \Cref{fig:image_vs_embedding_s1,fig:image_vs_embedding_s2}. The forecasted images appear close to the targets in texture and color, but once converted into the embedding space, much of the high-level information is lost or altered. In S1\_Embed, the converted embeddings collapse to a few dominant colors. In S2\_Embed, their
PCA maps even resemble the target, but the error maps on all dimensions
show large differences in the other principal components, while the error of
\method{} remains close to zero.

\begin{figure*}[t]
  \centering
  \includegraphics[width=\linewidth]{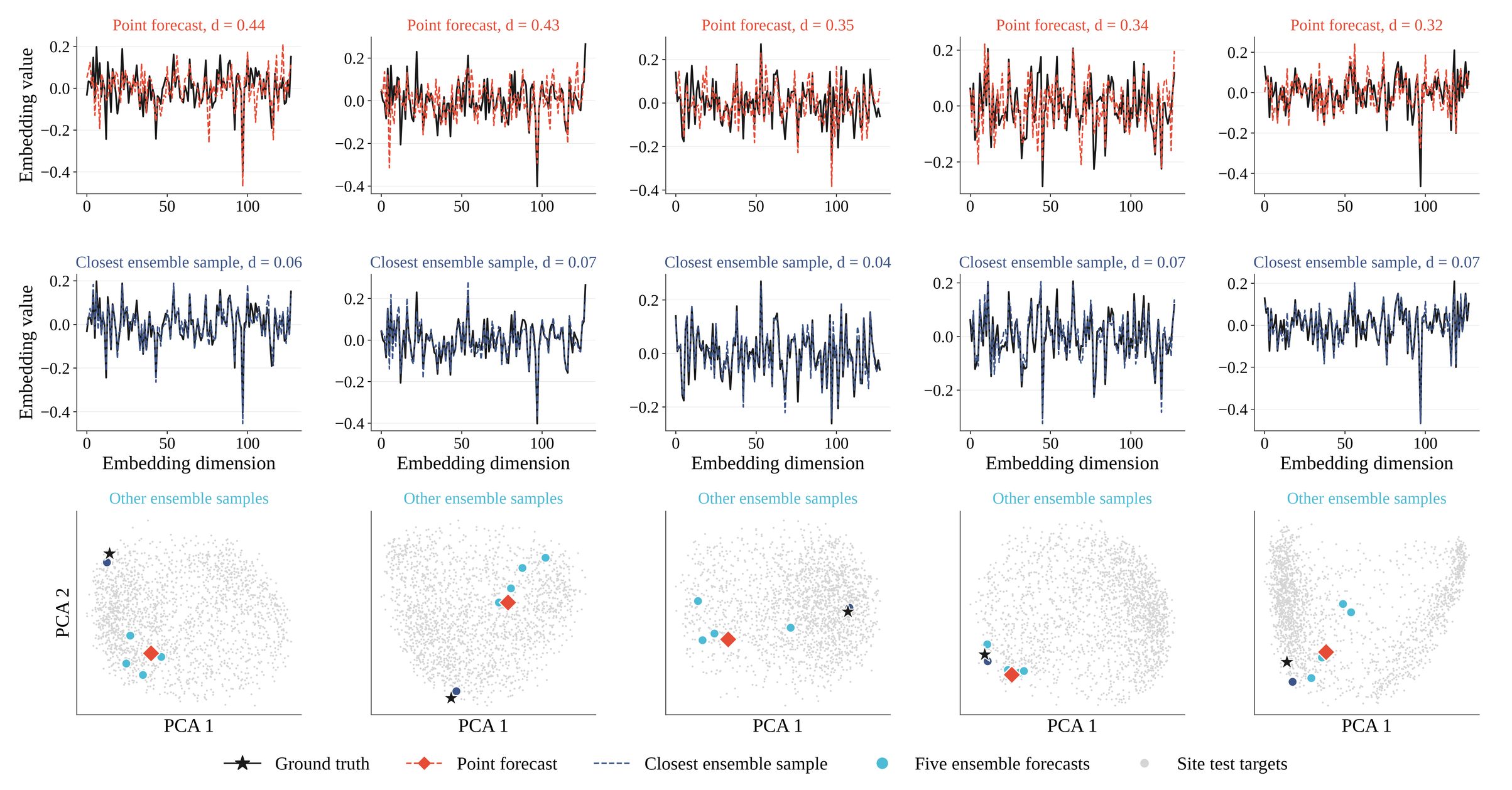}
  \caption{\textbf{Point \vs ensemble forecasts of \method{} on S1\_Embed.} Each column is a
  random test sample. Top: the point forecast against the ground truth over the
  128 dimensions. Middle: the ensemble member closest to the ground truth.
  Bottom: five ensemble members projected onto the first two PCA components
  of the test targets of the same site. $d$ is the cosine distance to the ground truth.}
  \label{fig:point_vs_ensemble_s1}
\end{figure*}
\paragraph{Learning normal change in the embedding space generalizes across anomaly categories.}
As shown in \cref{tab:gsa}, \method{} achieves the best average precision and
F$_1$, raising F$_1$ from 41.58 of ESIA to 48.70, and it gives the best F$_1$
in five of the six categories. The time-series detectors COLD and PKAD obtain
higher average recall, but their precision stays below 30, \ie they flag much
of the normal variation as anomalies. Spectral indices are strong only on the
category they were designed for, such as dNDVI on burned area. \method{}
instead learns what normal change looks like in the embedding space, so the
same model works across all categories without category-specific design.
Marine debris is the exception, where \method{} ranks third in F$_1$: the
debris covers only a few pixels, so even small false alarms, such as the
scattered high scores along the coastline in \cref{fig:gsa_qualitative},
noticeably lower its precision. In the other categories, the forecast
embedding resembles the observed one everywhere except in the event region, so
the anomaly score concentrates on the annotated anomalies.

\subsection{Discussion}
\label{sec:exp_analysis}
We now place our results in a broader context and quantify the benefits from our design choices. 

\paragraph{What benefit does an ensemble forecast bring?}
Unlike deterministic models, \method{} generates ensemble forecasts to account for
factors that the context cannot predict. However, \textit{more} forecasts do not
necessarily mean a \textit{better} forecast, as the probabilistic DeepAR and TFT in
\cref{tab:main} show. In a low-dimensional space, such as the two VV and VH
bands of an S1 image, drawing 100 samples from the predicted Gaussian does
improve DeepAR's point forecast--from 0.1706 to 0.0926--and TFT's point 
forecast--from 0.1698 to 0.1039--in cosine distance. 
However, as the dimensionality increases, such as to the ten bands of an S2
image, the same sampling worsens the forecast, from 0.1812 to 0.2287 (DeepAR) and from
0.2359 to 0.2539 (TFT). Indeed, in 
the 128- and 1024-dimensional embedding spaces, the best
of 100 Gaussian samples is \textit{worse} than the point forecast of the same model on
both datasets. We hypothesize that the reason for this effect
is that, assuming dimensions are independent, a sample drifts further from the mean as
the dimension grows, so both DeepAR and TFT struggled to generate realistic embeddings via direct sampling.

Generative models
avoid this issue by producing whole embeddings without a distribution assumption, 
but, nevertheless, not all of them succeed.  For example,
TMDM$^*$ appears to predicts the noise with its ensemble cosine distance
rising from 0.1210 to 0.3211 and from 0.0948 to 0.3359 (denoising models of
this type are known to struggle with high-dimensional
signals~\cite{li2025jit} ) .  Instead, SimDiff$^*$ and \method{}  follow the
$x$-prediction style that outputs the embedding directly rather than noises \cite{li2025jit}, and both improve performance
from an ensemble with only 10 members: SimDiff$^*$ from 0.1105 to 0.1019 on S1\_Embed, and \method{} from 0.1087 to 0.0911.

To show this benefit more clearly, we visualize the forecasts of \method{} for
random test samples of S1\_Embed in \cref{fig:point_vs_ensemble_s1} (the corresponding S2\_Embed in \cref{fig:point_vs_ensemble_s2} of \cref{sec:app_point_vs_ensemble}). The point
forecast follows the overall trend of the target but misses many individual
dimensions, whereas the closest ensemble member reproduces almost every
dimension, reducing $d$ from 0.32--0.44 to 0.04--0.07 on S1\_Embed and from
0.58--0.65 to 0.16 on S2\_Embed. Projected into 2D, the five ensemble members

spread around the point forecast, which sits near their centre, and form a
local manifold of plausible futures.

\begin{table}[t]
  \centering
  \small
  \setlength{\tabcolsep}{4pt}
  \begin{tabular}{@{}l|ccccc@{}}
    \toprule
    Historical number & 3 & 5 & 7 & 9 & 12 \\
    \midrule
    Std $z$ $\rightarrow1$       & 1.198 & 1.162 & 1.138 & 1.131 & 1.134 \\
    Ensemble Cosine $\downarrow$ & 0.1007 & 0.0945 & 0.0911 & 0.0902 & 0.0898 \\
    \bottomrule
  \end{tabular}
  \caption{\textbf{Effect of the historical number on S1\_Embed.}}
  \label{tab:history}
\end{table}

\paragraph{What if longer historical observations are provided?}
We train and test \method{} on S1\_Embed with 3, 5, 7, 9 and 12 historical
observations per series. 
As before, every series has its own irregular
intervals and lead time. As \cref{tab:history} shows, a longer history gives a
more accurate estimate of the uncertainty: std $z$ falls stably. It is therefore not surprising that S2\_Embed, with 37
historical observations, reaches a std $z$ of 1.0250 (\cref{tab:main}). Besides, more
prior information does not suppress the generative ability, rather, it makes
the ensemble more realistic, and the ensemble cosine distance decreases from
0.1007 to 0.0898. We find that beyond 7 observations both
metrics change only marginally, so we use 7 historical observations for S1\_Embed.

\begin{table}[!t]
  \centering
  \footnotesize
  \setlength{\tabcolsep}{2.5pt}
  \newcommand{\hd}[1]{\begin{tabular}[c]{@{}c@{}}#1\end{tabular}}
  \begin{tabular}{@{}l|ccccc@{}}
    \toprule
    Setting
      & \hd{Mean $z$\\$\rightarrow0$} & \hd{Std $z$\\$\rightarrow1$} & \hd{Point\\Cosine $\downarrow$} & \hd{Ensemble\\Cosine $\downarrow$} & CRPS $\downarrow$ \\
    \midrule
    w/o date token & $\mathbf{-0.0001}$ & 1.2324 & 0.1188 & 0.1008 & 0.0224 \\
    w/o location & 0.0012 & 1.1471 & 0.1100 & 0.0924 & 0.0214 \\
    \midrule
    SH~\cite{klemmer2024satclip}  & 0.0006 & 1.1627 & \textbf{0.1078} & 0.0916 & \textbf{0.0213} \\
    SHDD~\cite{wu2025locdiff}     & 0.0007 & 1.1483 & 0.1088 & 0.0924 & 0.0216 \\
    RFF~\cite{cepeda2023geoclip} & $-0.0005$ & \textbf{1.1376} & 0.1087 & \textbf{0.0911} & \textbf{0.0213} \\
    \bottomrule
  \end{tabular}
  \caption{\textbf{Ablation of the date and location tokens on S1\_Embed.}
  ``w/o date token'' removes the date encoding added to the historical
  observation tokens (\cref{eq:obstoken}); The bottom block compares location encoders.}
  \label{tab:token_ablation}
\end{table}

\paragraph{What is the effect of date and location tokens?}
Recall that S1\_Embed covers sites across the globe.
To study the effect of location and date, we ablate the corresponding tokens. The date token tested here is the date encoding added to
each historical observation token in \cref{eq:obstoken}. Note that the \textit{query} date token
is necessary for the forecast and is therefore always kept. Besides removing the
location token, we also replace RFF with other location encoders: the
spherical harmonics (SH) of SatCLIP~\cite{klemmer2024satclip} and the
spherical harmonics with discretized-DFS (SHDD) of LocDiffusion~\cite{wu2025locdiff}. As
\cref{tab:token_ablation} shows, removing either token hurts the performance. Removing the date token matters more: it is the only setting
with a std $z$ above 1.2 (1.2324) and an ensemble cosine distance above 0.10
(0.1008). Among the location
encoders, SH and RFF perform better than SHDD. Considering the large difference in std $z$, 1.1382 against
1.1627, we choose RFF as our location encoder.

\paragraph{What is the effect of the forecast lead time?}
We group the test samples of both forecast datasets by the lead time from the last
historical observation to the target date and evaluate each group separately
(\cref{fig:lead_time} in \cref{sec:app_lead_time}). On both datasets, the uncertainty estimate remains
stable as the lead time grows: mean $z$ stays within $\pm$0.0015, and std $z$
stays between 1.12 and 1.21 on S1\_Embed and close to 1 on S2\_Embed (\cref{fig:lead_time}a,d). The forecast itself, however, gradually
degrades with lead times. Both the point and the ensemble cosine
distance rise (\cref{fig:lead_time}b,e), as does the CRPS
(\cref{fig:lead_time}c,f); on S2\_Embed the CRPS almost doubles, from 0.0033
at 5 days to 0.0060 at 37 days. One exception is the 11-day group of
S1\_Embed, which is even better than the 4-day group, with a point cosine
distance of 0.1045 against 0.1125. We attribute this to its size: with 532k
items, it holds several times more samples than other groups, so the model
is most familiar with this lead time.

\begin{figure}[t]
  \centering
  \includegraphics[width=\linewidth]{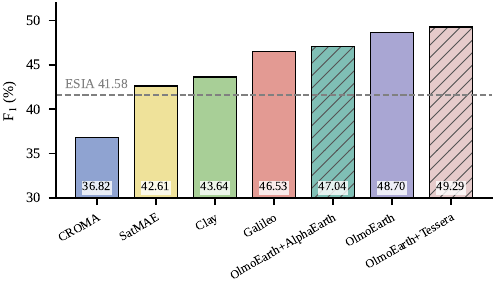}
  \caption{\textbf{Average F$_1$ of \method{} on GSA with different embedding
  spaces.} Hatched bars add annual Tessera or AlphaEarth embeddings as extra
  conditions to OlmoEarth, and the dashed line marks ESIA, the best baseline in
  \cref{tab:gsa}. Recall and precision are given in \cref{tab:gsa_embedding}.}
  \label{fig:gsa_embedding}
\end{figure}

\begin{table*}[!t]
  \centering
  \small
  \setlength{\tabcolsep}{3.5pt}
  \begin{tabular}{@{}lccccccccc@{\hspace{10pt}}c@{}}
    \toprule
    & & & & & \multicolumn{4}{c}{SatCLIP~\cite{klemmer2024satclip}} & & \\
    \cmidrule(lr){6-9}
    Task & Identity & GPS2Vec & CSP-iNat & CSP-FMoW
         & \shortstack{RN18\\$L{=}10$} & \shortstack{RN18\\$L{=}40$}
         & \shortstack{ViT-16\\$L{=}10$} & \shortstack{ViT-16\\$L{=}40$}
         & GeoCLIP & \method{} \\
    \midrule
    Population    & $-2.9862$  & $-0.5996$ & $-2.0198$ & $-0.1886$ & $0.1468$  & $0.2373$  & $-0.1801$ & $0.1028$  & $0.3717$          & $\mathbf{0.6064}$ \\
    Elevation     & $-6.8305$  & $-0.1472$ & $-4.4808$ & $-0.5511$ & $-19.2185$& $-3.6502$ & $-3.4695$ & $-0.0937$ & $-1.9232$         & $\mathbf{0.1687}$ \\
    Temperature   & $-94.6671$ & $-0.2646$ & $-0.6582$ & $0.1980$  & $0.2065$  & $0.5840$  & $0.3687$  & $0.5450$  & $0.7094$          & $\mathbf{0.7628}$ \\
    House value   & $-0.0118$  & $-0.0316$ & $0.0774$  & $-0.5702$ & $0.2182$  & $-0.2985$ & $0.0083$  & $-2.8809$ & $\mathbf{0.4740}$ & $0.2183$ \\
    Median income & $-0.0185$  & $0.5964$  & $0.1035$  & $-1.8341$ & $0.1644$  & $0.0224$  & $0.5032$  & $0.2731$  & $\mathbf{0.6646}$ & $0.5282$ \\
    \midrule
    Average       & $-20.9028$ & $-0.0893$ & $-1.3956$ & $-0.5892$ & $-3.6965$ & $-0.6210$ & $-0.5539$ & $-0.4107$ & $0.0593$          & $\mathbf{0.4569}$ \\
    \bottomrule
  \end{tabular}
  \caption{\textbf{Results when adapting \method{} as INR on five regression tasks ($R^{2}$).}
  All methods share the same probe and spatial splits. $L$: Legendre degree of
  SatCLIP. Best per row in bold.}
  \label{tab:spatial}
\end{table*}
\begin{figure*}[t]
  \centering
  \includegraphics[width=0.95\textwidth]{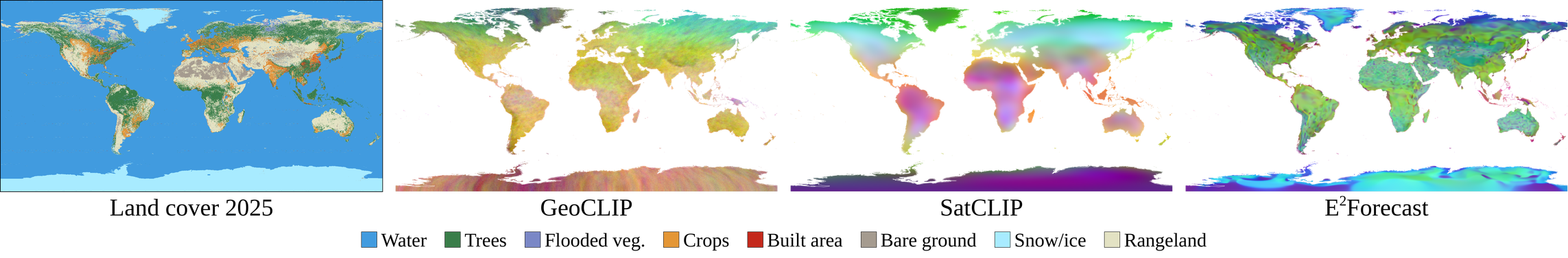}
  \caption{\textbf{The global embedding field compared with different INR models.} Embeddings queried on a
  $0.5^{\circ}$ grid, projected to RGB by PCA, with the ocean
  masked. Left: the 2025 Esri land-cover map~\cite{karra2021lulc}.}
  \label{fig:global_field}
\end{figure*}

\begin{figure*}[t]
  \centering
  \includegraphics[width=\textwidth]{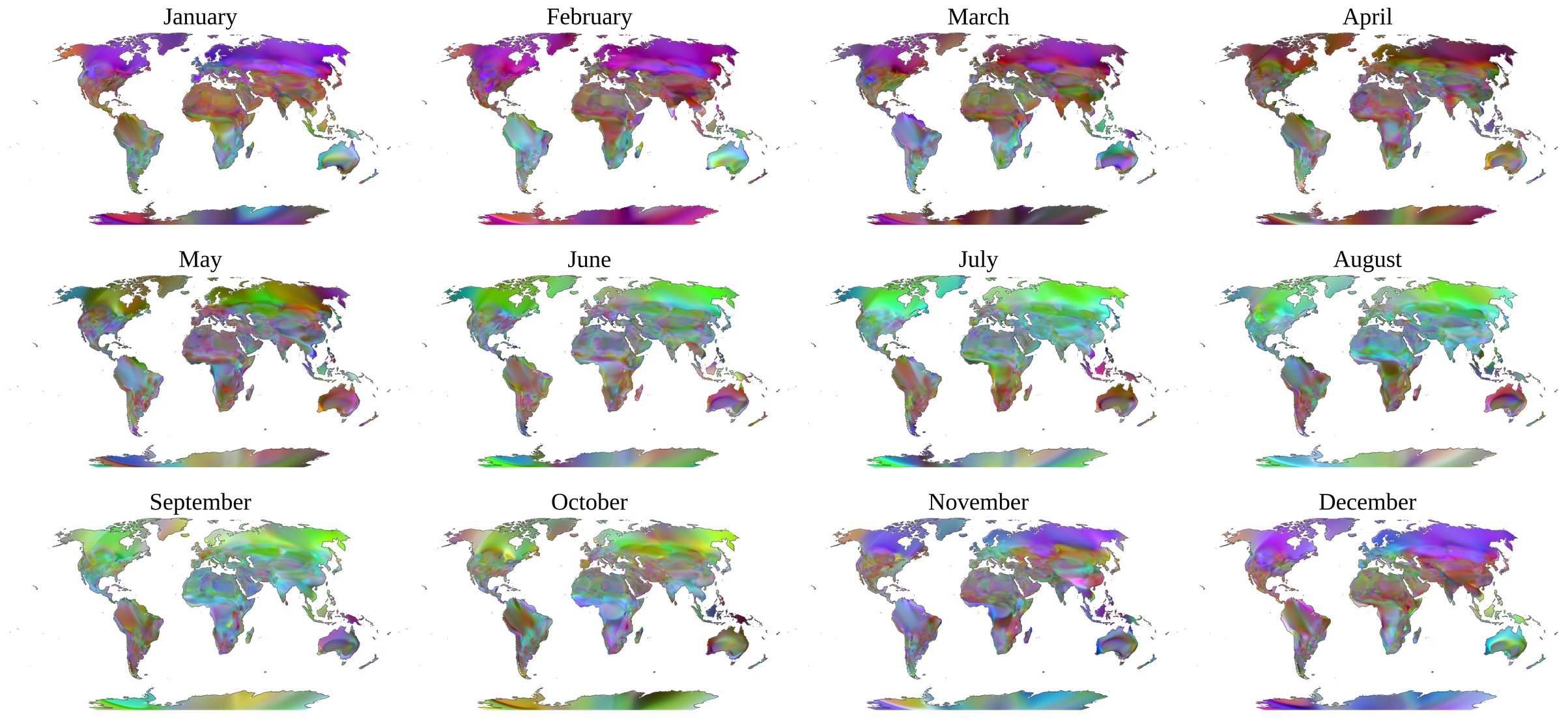}
  \caption{\textbf{Time- and location-aware global embedding field.} \method{} trained on
  SSL4EO-S12~\cite{wang2023ssl4eo} with the query date token, queried on a $0.5^{\circ}$ grid for
  each month. The annual mean is subtracted at each location before projecting to RGB.}
  \label{fig:global_monthly}
\end{figure*}

\paragraph{Is \method{} compatible with different embedding spaces?}
We have already shown that \method{} works in the OlmoEarth and Prithvi
embedding spaces on S1\_Embed and S2\_Embed. We additionally test seven
embedding spaces on GSA (\cref{fig:gsa_embedding}; full results in
\cref{tab:gsa_embedding} of \cref{sec:app_gsa_embedding}): OlmoEarth~\cite{herzog2025olmoearth},
Galileo~\cite{tseng2025galileo}, CROMA~\cite{fuller2023croma},
SatMAE~\cite{cong2022satmae} and Clay~\cite{clay2024}, and OlmoEarth combined
with the annual Tessera or AlphaEarth embeddings. 
Since anomalies are annotated per pixel, we only use foundation models that support a
patch size no larger than 8, which leaves out those with a patch size of 16, such
as Prithvi-EO-2.0 \cite{szwarcman2024prithvi} and DOFA~\cite{xiong2024dofa}. With the same model and
training recipe, six of the seven spaces exceed the best baseline ESIA \cite{li2026esia}
(41.58 F$_1$), showing that \method{} does not depend on a particular embedding space.
In general, a stronger foundation model brings better
results. For example, the recent multi-temporal, multi-sensor OlmoEarth and Galileo reach
48.70 and 46.53 F$_1$, ahead of the older Clay, SatMAE and CROMA  models (43.64, 42.61 and
36.82). Besides, different embedding spaces can also be used together. We find that adding Tessera v1.1
to OlmoEarth gives the best result, raising precision to 49.65 and F$_1$ to
49.29.

\section{Connection with implicit neural representations}
\label{sec:exp_spatial}

In this section, we explore the connection between implicit neural representations and \method{}.
An implicit neural representation (INR) stores a signal as a network that maps a
coordinate to the value there \cite{sitzmann2020siren}. With an empty context ($\mathcal{C}=\emptyset$
in \cref{eq:instance}), \method{} maps the location token
alone to an embedding at $(\lambda,\varphi)$, \ie it becomes an INR of the global
field of foundation-model embeddings.

To test the performance of such an INR, we compare \method{} with an empty context with
SatCLIP~\cite{klemmer2024satclip}.
To do a fair comparison, we train on the same $10^{5}$ Sentinel-2 embeddings as SatCLIP.
At the inference stage, we used our trained INR model to generate semantic embeddings for each unseen location, and then tuned an MLP layer to regress it onto the five tasks of
population density, elevation, air temperature~\cite{klemmer2024satclip,hooker2018temperature} and California
median house value and income~\cite{pace1997housing}, reporting the test $R^{2}$.
For the two California Housing tasks, five z-scored tabular features (HouseAge, AveRooms, AveBedrms, Population, AveOccup) are appended to the embedding for all methods.
Since SatCLIP does not
release its downstream splits, and a random split may leak information, we therefore split each dataset into disjoint train/val/test
regions ($30/10/60$ bands along latitude or longitude, with a $0.25^{\circ}$ guard gap, as shown as the
top row of \cref{fig:qualitative} in \cref{sec:app_qualitative}).

We compare our approach with with SatCLIP (all four released checkpoints, ResNet-18 or ViT-16 with degree $L\!=\!10$ or $40$), as well as GeoCLIP,
CSP~\cite{mai2023csp}, GPS2Vec~\cite{yin2019gps2vec} and raw coordinates
(\cref{tab:spatial}). Under region-disjoint splits, every deterministic encoder exhibited degraded performance, with eight of nine baselines have a negative average $R^{2}$.
The best,
GeoCLIP, reaches an average $R^{2}$ of only $0.0593$, because the probe extrapolates a fixed spherical
basis rather than the target.
\method{} is the only method positive on all five tasks and raises the average $R^{2}$
to $0.4569$. GeoCLIP leads only on the two California tasks,
where
the geographic shift between the train and test spans fewer degrees than the other tasks. In \cref{fig:qualitative} (\cref{sec:app_qualitative}), \method{} reproduces the large-scale spatial patterns of population, relief, and temperature in the held-out regions.

For a visual comparison, we show the generated global-scale embeddings in \cref{fig:global_field}.
Each encoder is queried on a
global $0.5^{\circ}$ grid, and embeddings are projected to RGB beside a 2025
land-cover map~\cite{karra2021lulc}. Although it never saw land-cover labels,
\method{} produces a field that follows coastlines and ice sheets and whose
regions match land cover: the Sahara and Arabian deserts, the Amazon and Congo forests, and the boreal belt are clearly visible.
In contrast, SatCLIP, which is aligned to the same embeddings using a contrastive loss, shows only a
smooth low-frequency function of the coordinate, which explains why its probe
does not extrapolate well. Similarly, GeoCLIP adds speckle and polar streaks unrelated to the surface.

Because the S2-100K dataset used in \cref{fig:global_field} has no acquisition dates, we further train \method{} with the same SatCLIP vision
encoder on the SSL4EO-S12 dataset~\cite{wang2023ssl4eo},
which samples four seasons per location, and enables the
query date token to generate a field for each month (\cref{fig:global_monthly}).
After removing the
annual mean, the field shows a clear yearly cycle: December and June are near opposite states, and the
Northern and Southern Hemispheres change in antiphase, \eg, boreal regions and Australia swap colors
between winter and summer. This strongly suggests that \method{} has learned both regional and seasonal structure.

\FloatBarrier %
\section{Conclusion}
\label{sec:conclusion}

We present \method{}  that makes forecasts in the embedding space of Earth
foundation models and find that it is more effective than 
prior approaches that forecast pixels in the native spectral space.
Its key innovation is a token space that allows it to take as input 
\textit{irregular} histories from \textit{different} sensors and foundation models and 
forecast at \textit{any} query date. 
Across Sentinel-1 and Sentinel-2
benchmarks, adapting existing forecasters into embedding space consistently
improves their accuracy and calibration as in Table \ref{tab:main}, with \method{} resulting in
the best point and ensemble forecasts.
Interestingly, the same forecasts also detect anomalous change without
labels or retraining. Moreover,
with an empty context, the model becomes a powerful implicit
neural representation of the global embedding field.

We acknowledge that \method{} still has two limitations. First, the
probability-estimation branch ignores the correlation between embedding
dimensions. This is because the surrogate in \cref{eq:surrogate} uses a diagonal covariance,
predicting the mean and standard deviation of each dimension independently,
whereas the dimensions of foundation-model embeddings are usually highly
correlated. We tried extending $\mathcal{L}_{\mathrm{nll}}$ to model this
cross-dimensional correlation, but it substantially degraded performance. We believe
this reflects our particular formulation rather than the idea itself, and leave
a better way of modelling these correlations to future work. Second, \method{}
forecasts only from past embeddings, location and date, without accounting for the external
factors that drive land surface change, such as weather. which govern vegetation growth,
floods and fires, and terrain, which shapes how water and land cover evolve. Adding
such Earth system drivers as extra condition tokens is a natural next step that we intend to explore in future work.

{
    \small
    \bibliographystyle{ieeenat_fullname}
    \bibliography{main}

\begin{thebibliography}{76}
\providecommand{\natexlab}[1]{#1}
\providecommand{\url}[1]{\texttt{#1}}
\expandafter\ifx\csname urlstyle\endcsname\relax
  \providecommand{\doi}[1]{doi: #1}\else
  \providecommand{\doi}{doi: \begingroup \urlstyle{rm}\Url}\fi

\bibitem[Alawadi(2010)]{alawadi2010sabi}
Fahad Alawadi.
\newblock Detection of surface algal blooms using the newly developed algorithm
  surface algal bloom index ({SABI}).
\newblock In \emph{Remote Sensing of the Ocean, Sea Ice, and Large Water
  Regions}, pages 45--58. SPIE, 2010.

\bibitem[Al{\`i} et~al.(2024)Al{\`i}, Razzano, Vitale, Ferraioli, Pascazio,
  Schirinzi, and Ullo]{ali2024fdi}
Mattia Al{\`i}, Francesco Razzano, Sergio Vitale, Giampaolo Ferraioli, Vito
  Pascazio, Gilda Schirinzi, and Silvia Ullo.
\newblock Assessment of spectral-based solutions for the detection of floating
  marine debris.
\newblock In \emph{International Conference on Networks, Multimedia and
  Information Technology (NMITCON)}, pages 1--5, 2024.

\bibitem[Benson et~al.(2024)Benson, Robin, Requena-Mesa, Alonso, Carvalhais,
  Cort{\'e}s, Gao, Linscheid, Weynants, and
  Reichstein]{benson2024contextformer}
Vitus Benson, Claire Robin, Christian Requena-Mesa, Lazaro Alonso, Nuno
  Carvalhais, Jos{\'e} Cort{\'e}s, Zhihan Gao, Nora Linscheid, M{\'e}lanie
  Weynants, and Markus Reichstein.
\newblock Multi-modal learning for geospatial vegetation forecasting.
\newblock In \emph{CVPR}, pages 27788--27799, 2024.

\bibitem[Bi et~al.(2023)Bi, Xie, Zhang, Chen, Gu, and Tian]{bi2023pangu}
Kaifeng Bi, Lingxi Xie, Hengheng Zhang, Xin Chen, Xiaotao Gu, and Qi Tian.
\newblock Accurate medium-range global weather forecasting with {3D} neural
  networks.
\newblock \emph{Nature}, 619:\penalty0 533--538, 2023.

\bibitem[Bovolo and Bruzzone(2007)]{bovolo2007cva}
Francesca Bovolo and Lorenzo Bruzzone.
\newblock A theoretical framework for unsupervised change detection based on
  change vector analysis in the polar domain.
\newblock \emph{IEEE Transactions on Geoscience and Remote Sensing},
  45\penalty0 (1):\penalty0 218--236, 2007.

\bibitem[Brekke and Solberg(2005)]{brekke2005ndoi}
Camilla Brekke and Anne H.~S. Solberg.
\newblock Oil spill detection by satellite remote sensing.
\newblock \emph{Remote Sensing of Environment}, 95:\penalty0 1--13, 2005.

\bibitem[Brown et~al.(2025)Brown, Kazmierski, Pasquarella, Rucklidge,
  Samsikova, Zhang, Shelhamer, Lahera, Wiles, Ilyushchenko, Gorelick, Zhang,
  Alj, Schechter, Askay, Guinan, Moore, Boukouvalas, and
  Kohli]{brown2025alphaearth}
Christopher~F. Brown, Michal~R. Kazmierski, Valerie~J. Pasquarella, William~J.
  Rucklidge, Masha Samsikova, Chenhui Zhang, Evan Shelhamer, Estefania Lahera,
  Olivia Wiles, Simon Ilyushchenko, Noel Gorelick, Lihui~Lydia Zhang, Sophia
  Alj, Emily Schechter, Sean Askay, Oliver Guinan, Rebecca Moore, Alexis
  Boukouvalas, and Pushmeet Kohli.
\newblock {AlphaEarth Foundations}: An embedding field model for accurate and
  efficient global mapping from sparse label data.
\newblock \emph{arXiv preprint arXiv:2507.22291}, 2025.

\bibitem[Castillo-Villamor et~al.(2021)Castillo-Villamor, Hardy, Bunting,
  Llanos-Peralta, Zamora, Rodriguez, and Gomez-Latorre]{castillo2021eoad}
Liliana Castillo-Villamor, Andy Hardy, Pete Bunting, Wilson Llanos-Peralta,
  Miguel Zamora, Yeison Rodriguez, and Diego~Alejandro Gomez-Latorre.
\newblock The {Earth Observation-based Anomaly Detection (EOAD)} system: A
  simple, scalable approach to mapping in-field and farm-scale anomalies using
  widely available satellite imagery.
\newblock \emph{International Journal of Applied Earth Observation and
  Geoinformation}, 104, 2021.

\bibitem[Cepeda et~al.(2023)Cepeda, Nayak, and Shah]{cepeda2023geoclip}
Vicente~Vivanco Cepeda, Gaurav~Kumar Nayak, and Mubarak Shah.
\newblock {GeoCLIP}: Clip-inspired alignment between locations and images for
  effective worldwide geo-localization.
\newblock In \emph{NeurIPS}, 2023.

\bibitem[Chen et~al.(2026)Chen, Wang, Yuan, Li, Xia, Liao, Wei, Yuan, Xu, Zhu,
  Zhu, Zhang, Zhou, Fu, Huang, Chen, Dai, and Gong]{chen2026esd}
Shuang Chen, Jie Wang, Shuai Yuan, Jiayang Li, Yu Xia, Yuanhong Liao, Junbo
  Wei, Jincheng Yuan, Xiaoqing Xu, Xiaolin Zhu, Peng Zhu, Hongsheng Zhang, Yuyu
  Zhou, Haohuan Fu, Huabing Huang, Bin Chen, Fan Dai, and Peng Gong.
\newblock Democratizing planetary-scale analysis: an ultra-lightweight {Earth}
  embedding database for accurate and flexible global land monitoring.
\newblock \emph{Earth System Science Data}, 18:\penalty0 5375--5398, 2026.

\bibitem[Cheng et~al.(2026)Cheng, Zhu, and Fan]{cheng2026landslide}
Yusen Cheng, Qinfeng Zhu, and Lei Fan.
\newblock From landslide conditioning factors to satellite embeddings:
  Evaluating the utilisation of {Google AlphaEarth} for landslide
  susceptibility mapping using deep learning.
\newblock \emph{arXiv preprint arXiv:2601.07268}, 2026.

\bibitem[{Clay Foundation}(2024)]{clay2024}
{Clay Foundation}.
\newblock {Clay}: A foundation model for {E}arth observation.
\newblock \url{https://github.com/Clay-foundation/model}, 2024.

\bibitem[Cong et~al.(2022)Cong, Khanna, Meng, Liu, Rozi, He, Burke, Lobell, and
  Ermon]{cong2022satmae}
Yezhen Cong, Samar Khanna, Chenlin Meng, Patrick Liu, Erik Rozi, Yutong He,
  Marshall Burke, David~B. Lobell, and Stefano Ermon.
\newblock {SatMAE}: Pre-training transformers for temporal and multi-spectral
  satellite imagery.
\newblock In \emph{NeurIPS}, 2022.

\bibitem[Czerkawski et~al.(2024)Czerkawski, Kluczek, and
  Bojanowski]{czerkawski2024majortom}
Mikolaj Czerkawski, Marcin Kluczek, and J\k{e}drzej~S. Bojanowski.
\newblock Global and dense embeddings of {E}arth: {Major TOM} floating in the
  latent space.
\newblock \emph{arXiv preprint arXiv:2412.05600}, 2024.

\bibitem[Ding et~al.(2026)Ding, Wang, Zhou, and Yao]{ding2026simdiff}
Hang Ding, Xue Wang, Tian Zhou, and Tao Yao.
\newblock {SimDiff}: Simpler yet better diffusion model for time series point
  forecasting.
\newblock \emph{Proceedings of the AAAI Conference on Artificial Intelligence},
  40\penalty0 (25):\penalty0 20781--20789, 2026.

\bibitem[Fang et~al.(2026)Fang, Stewart, Corley, Zhu, and
  Azizpour]{fang2026products}
Heng Fang, Adam~J. Stewart, Isaac Corley, Xiao~Xiang Zhu, and Hossein Azizpour.
\newblock {E}arth embeddings as products: Taxonomy, ecosystem, and standardized
  access.
\newblock \emph{arXiv preprint arXiv:2601.13134}, 2026.

\bibitem[Feng et~al.(2026{\natexlab{a}})Feng, Atzberger, Jaffer, Knezevic,
  Sormunen, Young, Lisaius, Immitzer, Jackson, Ball, Coomes, Madhavapeddy,
  Blake, and Keshav]{feng2026tessera}
Zhengpeng Feng, Clement Atzberger, Sadiq Jaffer, Jovana Knezevic, Silja
  Sormunen, Robin Young, Madeline~C. Lisaius, Markus Immitzer, Toby Jackson,
  James Ball, David~A. Coomes, Anil Madhavapeddy, Andrew Blake, and Srinivasan
  Keshav.
\newblock {TESSERA}: Temporal embeddings of surface spectra for {E}arth
  representation and analysis.
\newblock In \emph{CVPR}, 2026{\natexlab{a}}.

\bibitem[Feng et~al.(2026{\natexlab{b}})Feng, Jaffer, Shokar, Knezevic, Ball,
  Sousa, Elvers, Lisaius, Atzberger, Young, et~al.]{feng2026tesserav2}
Zhengpeng Feng, Sadiq Jaffer, Ira Shokar, Jovana Knezevic, James Ball, Pedro
  Sousa, Mark Elvers, Madeline Lisaius, Clement Atzberger, Robin Young, et~al.
\newblock Tessera v2: Scaling pixel-wise earth foundation models.
\newblock \emph{arXiv preprint arXiv:2607.03949}, 2026{\natexlab{b}}.

\bibitem[Fuller et~al.(2023)Fuller, Millard, and Green]{fuller2023croma}
Anthony Fuller, Koreen Millard, and James~R. Green.
\newblock {CROMA}: Remote sensing representations with contrastive
  radar-optical masked autoencoders.
\newblock In \emph{NeurIPS}, 2023.

\bibitem[Gao et~al.(2022)Gao, Shi, Wang, Zhu, Wang, Li, and
  Yeung]{gao2022earthformer}
Zhihan Gao, Xingjian Shi, Hao Wang, Yi Zhu, Yuyang Wang, Mu Li, and Dit-Yan
  Yeung.
\newblock {Earthformer}: Exploring space-time transformers for {E}arth system
  forecasting.
\newblock In \emph{NeurIPS}, 2022.

\bibitem[Garnot and Landrieu(2021)]{garnot2021panoptic}
Vivien Sainte~Fare Garnot and Loic Landrieu.
\newblock Panoptic segmentation of satellite image time series with
  convolutional temporal attention networks.
\newblock In \emph{ICCV}, pages 4872--4881, 2021.

\bibitem[Gneiting and Raftery(2007)]{gneiting2007strictly}
Tilmann Gneiting and Adrian~E. Raftery.
\newblock Strictly proper scoring rules, prediction, and estimation.
\newblock \emph{Journal of the American Statistical Association}, 102\penalty0
  (477):\penalty0 359--378, 2007.

\bibitem[Gu et~al.(2025)Gu, Mao, and Shou]{gu2025long}
Yuchao Gu, Weijia Mao, and Mike~Zheng Shou.
\newblock Long-context autoregressive video modeling with next-frame
  prediction.
\newblock \emph{arXiv preprint arXiv:2503.19325}, 2025.

\bibitem[Herzog et~al.(2025)Herzog, Bastani, Zhang, Tseng, Redmon, Sablon,
  Park, Morrison, et~al.]{herzog2025olmoearth}
Henry Herzog, Favyen Bastani, Yawen Zhang, Gabriel Tseng, Joseph Redmon,
  Hadrien Sablon, Ryan Park, Jacob Morrison, et~al.
\newblock {OlmoEarth}: Stable latent image modeling for multimodal {E}arth
  observation.
\newblock \emph{arXiv preprint arXiv:2511.13655}, 2025.

\bibitem[Hooker et~al.(2018)Hooker, Duveiller, and
  Cescatti]{hooker2018temperature}
Josh Hooker, Gregory Duveiller, and Alessandro Cescatti.
\newblock A global dataset of air temperature derived from satellite remote
  sensing and weather stations.
\newblock \emph{Scientific Data}, 5:\penalty0 180246, 2018.

\bibitem[Hu(2009)]{hu2009fai}
Chuanmin Hu.
\newblock A novel ocean color index to detect floating algae in the global
  oceans.
\newblock \emph{Remote Sensing of Environment}, 113:\penalty0 2118--2129, 2009.

\bibitem[Hu et~al.(2021)Hu, Ban, and Nascetti]{hu2021afds2}
Xikun Hu, Yifang Ban, and Andrea Nascetti.
\newblock {Sentinel-2 MSI} data for active fire detection in major fire-prone
  biomes: A multi-criteria approach.
\newblock \emph{International Journal of Applied Earth Observation and
  Geoinformation}, 101:\penalty0 102347, 2021.

\bibitem[Jiang et~al.(2014)Jiang, Feng, Zhu, Lu, Huang, and
  Xiao]{jiang2014awei}
Hao Jiang, Min Feng, Yunqiang Zhu, Ning Lu, Jianxi Huang, and Tong Xiao.
\newblock An automated method for extracting rivers and lakes from {Landsat}
  imagery.
\newblock \emph{Remote Sensing}, 6:\penalty0 5067--5089, 2014.

\bibitem[Jiang et~al.(2006)Jiang, Huete, Chen, Chen, Li, Yan, and
  Zhang]{jiang2006ndvi}
Zhangyan Jiang, Alfredo~R. Huete, Jin Chen, Yunhao Chen, Jing Li, Guangjian
  Yan, and Xiaoyu Zhang.
\newblock Analysis of {NDVI} and scaled difference vegetation index retrievals
  of vegetation fraction.
\newblock \emph{Remote Sensing of Environment}, 101:\penalty0 366--378, 2006.

\bibitem[Karra et~al.(2021)Karra, Kontgis, Statman-Weil, Mazzariello, Mathis,
  and Brumby]{karra2021lulc}
Krishna Karra, Caitlin Kontgis, Zoe Statman-Weil, Joseph~C. Mazzariello, Mark
  Mathis, and Steven~P. Brumby.
\newblock Global land use/land cover with {Sentinel 2} and deep learning.
\newblock In \emph{IEEE International Geoscience and Remote Sensing Symposium
  (IGARSS)}, pages 4704--4707, 2021.

\bibitem[Khanna et~al.(2024)Khanna, Liu, Zhou, Meng, Rombach, Burke, Lobell,
  and Ermon]{khanna2024diffusionsat}
Samar Khanna, Patrick Liu, Linqi Zhou, Chenlin Meng, Robin Rombach, Marshall
  Burke, David~B. Lobell, and Stefano Ermon.
\newblock {DiffusionSat}: A generative foundation model for satellite imagery.
\newblock In \emph{ICLR}, 2024.

\bibitem[Klemmer et~al.(2025)Klemmer, Rolf, Robinson, Mackey, and
  Ru{\ss}wurm]{klemmer2024satclip}
Konstantin Klemmer, Esther Rolf, Caleb Robinson, Lester Mackey, and Marc
  Ru{\ss}wurm.
\newblock {SatCLIP}: Global, general-purpose location embeddings with satellite
  imagery.
\newblock In \emph{AAAI}, 2025.

\bibitem[Kollovieh et~al.(2023)Kollovieh, Ansari, Bohlke-Schneider, Zschiegner,
  Wang, and Wang]{kollovieh2023predict}
Marcel Kollovieh, Abdul~Fatir Ansari, Michael Bohlke-Schneider, Jasper
  Zschiegner, Hao Wang, and Yuyang Wang.
\newblock Predict, refine, synthesize: Self-guiding diffusion models for
  probabilistic time series forecasting.
\newblock In \emph{Advances in Neural Information Processing Systems}, pages
  28341--28364, 2023.

\bibitem[Li et~al.(2025)Li, Liu, Wang, Peng, Sun, Wang, Sun, Ke, Jiang, Lu,
  Zhao, and Zhong]{li2025hyperfree}
Jingtao Li, Yingyi Liu, Xinyu Wang, Yunning Peng, Chen Sun, Shaoyu Wang,
  Zhendong Sun, Tian Ke, Xiao Jiang, Tangwei Lu, Anran Zhao, and Yanfei Zhong.
\newblock {HyperFree}: A channel-adaptive and tuning-free foundation model for
  hyperspectral remote sensing imagery.
\newblock In \emph{CVPR}, 2025.

\bibitem[Li et~al.(2026{\natexlab{a}})Li, Zhong, Zhu, Wang, and
  Zhang]{li2026anomalycd}
Jingtao Li, Yanfei Zhong, Qian Zhu, Xinyu Wang, and Liangpei Zhang.
\newblock Unsupervised anomaly localization with high-resolution and
  time-series satellite imagery: From global disaster dataset to zero-shot
  model via {SAM}.
\newblock \emph{ISPRS Journal of Photogrammetry and Remote Sensing},
  237:\penalty0 467--491, 2026{\natexlab{a}}.

\bibitem[Li et~al.(2026{\natexlab{b}})Li, Zhu, Wang, Li, Zhang, and
  Zhong]{li2026esia}
Jingtao Li, Qian Zhu, Xinyu Wang, Deren Li, Liangpei Zhang, and Yanfei Zhong.
\newblock Earth surface immune system for rapid monitoring of unknown
  anomalies.
\newblock \emph{arXiv preprint arXiv:2609.20662}, 2026{\natexlab{b}}.

\bibitem[Li and He(2025)]{li2025jit}
Tianhong Li and Kaiming He.
\newblock Back to basics: Let denoising generative models denoise.
\newblock \emph{arXiv preprint arXiv:2511.13720}, 2025.

\bibitem[Li et~al.(2024)Li, Chen, Hu, Chen, Sun, and Zhou]{li2024transformer}
Yuxin Li, Wenchao Chen, Xinyue Hu, Bo Chen, Baolin Sun, and Mingyuan Zhou.
\newblock Transformer-modulated diffusion models for probabilistic multivariate
  time series forecasting.
\newblock In \emph{International Conference on Learning Representations}, 2024.

\bibitem[Lim et~al.(2021)Lim, Ar{\i}k, Loeff, and Pfister]{lim2021temporal}
Bryan Lim, Sercan~{\"O}. Ar{\i}k, Nicolas Loeff, and Tomas Pfister.
\newblock Temporal fusion transformers for interpretable multi-horizon time
  series forecasting.
\newblock \emph{International Journal of Forecasting}, 37\penalty0
  (4):\penalty0 1748--1764, 2021.

\bibitem[Lipman et~al.(2023)Lipman, Chen, Ben-Hamu, Nickel, and
  Le]{lipman2023flowmatching}
Yaron Lipman, Ricky T.~Q. Chen, Heli Ben-Hamu, Maximilian Nickel, and Matt Le.
\newblock Flow matching for generative modeling.
\newblock In \emph{ICLR}, 2023.

\bibitem[Liu et~al.(2026)Liu, Zhang, Yang, and Zhu]{liu2026semantic}
Hao Liu, Xiaoxing Zhang, Chun Yang, and Xiaobin Zhu.
\newblock Semantic-enhanced time-series forecasting via large language models.
\newblock In \emph{International Conference on Learning Representations}, 2026.

\bibitem[Liu et~al.(2025{\natexlab{a}})Liu, Zhao, Wang, Wang, Du, Dang, and
  Mallorqui]{liu2025ndfi}
Jinglong Liu, Feng Zhao, Yunjia Wang, Yanan Wang, Sen Du, Lin Dang, and
  Jordi~J. Mallorqui.
\newblock Advancing coal fire detection model for large-scale areas based on
  {RS} indices and machine learning.
\newblock \emph{International Journal of Applied Earth Observation and
  Geoinformation}, 140:\penalty0 104587, 2025{\natexlab{a}}.

\bibitem[Liu et~al.(2023)Liu, Gong, and Liu]{liu2023rectified}
Xingchao Liu, Chengyue Gong, and Qiang Liu.
\newblock Flow straight and fast: Learning to generate and transfer data with
  rectified flow.
\newblock In \emph{ICLR}, 2023.

\bibitem[Liu et~al.(2025{\natexlab{b}})Liu, Qin, Huang, Wang, and
  Long]{liu2025timerxl}
Yong Liu, Guo Qin, Xiangdong Huang, Jianmin Wang, and Mingsheng Long.
\newblock {Timer-XL}: Long-context transformers for unified time series
  forecasting.
\newblock In \emph{International Conference on Learning Representations},
  2025{\natexlab{b}}.

\bibitem[Lutes et~al.(2006)Lutes, Keane, Caratti, Key, Benson, Sutherland, and
  Gangi]{lutes2006firemon}
Duncan~C. Lutes, Robert~E. Keane, John~F. Caratti, Carl~H. Key, Nathan~C.
  Benson, Steve Sutherland, and Larry~J. Gangi.
\newblock {FIREMON}: Fire effects monitoring and inventory system.
\newblock Technical Report RMRS-GTR-164, U.S. Department of Agriculture, Forest
  Service, Rocky Mountain Research Station, 2006.

\bibitem[Ma et~al.(2026)Ma, Shen, Swatantran, and Lobell]{ma2026harvesting}
Yuchi Ma, Yawen Shen, Anu Swatantran, and David~B. Lobell.
\newblock Harvesting {AlphaEarth}: Benchmarking the geospatial foundation model
  for agricultural downstream tasks.
\newblock \emph{arXiv preprint arXiv:2601.00857}, 2026.

\bibitem[Mai et~al.(2023)Mai, Lao, He, Song, and Ermon]{mai2023csp}
Gengchen Mai, Ni Lao, Yutong He, Jiaming Song, and Stefano Ermon.
\newblock {CSP}: Self-supervised contrastive spatial pre-training for
  geospatial-visual representations.
\newblock In \emph{Int. Conf. Mach. Learn.}, 2023.

\bibitem[Miller and Thode(2007)]{miller2007rdnbr}
Jay~D. Miller and Andrea~E. Thode.
\newblock Quantifying burn severity in a heterogeneous landscape with a
  relative version of the delta normalized burn ratio ({dNBR}).
\newblock \emph{Remote Sensing of Environment}, 109:\penalty0 66--80, 2007.

\bibitem[Nielsen(2007)]{nielsen2007irmad}
Allan~Aasbjerg Nielsen.
\newblock The regularized iteratively reweighted {MAD} method for change
  detection in multi- and hyperspectral data.
\newblock \emph{IEEE Transactions on Image Processing}, 16\penalty0
  (2):\penalty0 463--478, 2007.

\bibitem[Pace and Barry(1997)]{pace1997housing}
R.~Kelley Pace and Ronald Barry.
\newblock Sparse spatial autoregressions.
\newblock \emph{Statistics \& Probability Letters}, 33\penalty0 (3):\penalty0
  291--297, 1997.

\bibitem[Pickens et~al.(2025)Pickens, Hansen, Song, Poulson, Komarova, Baggett,
  Kerr, Mikus, {Ortiz Dominguez}, Tyukavina, and Lima]{pickens2025rapid}
Amy~H. Pickens, Matthew~C. Hansen, Zhen Song, Andrew Poulson, Anna Komarova,
  Antoine Baggett, Theodore Kerr, Aleksandra Mikus, Carolina {Ortiz Dominguez},
  Alexandra Tyukavina, and Andr{\'e} Lima.
\newblock Rapid monitoring of global land change.
\newblock \emph{Nature Communications}, 16\penalty0 (1), 2025.

\bibitem[Reed et~al.(2023)Reed, Gupta, Li, Brockman, Funk, Clipp, Keutzer,
  Candido, Uyttendaele, and Darrell]{reed2023scalemae}
Colorado~J. Reed, Ritwik Gupta, Shufan Li, Sarah Brockman, Christopher Funk,
  Brian Clipp, Kurt Keutzer, Salvatore Candido, Matt Uyttendaele, and Trevor
  Darrell.
\newblock {Scale-MAE}: A scale-aware masked autoencoder for multiscale
  geospatial representation learning.
\newblock In \emph{ICCV}, 2023.

\bibitem[Requena-Mesa et~al.(2021)Requena-Mesa, Benson, Reichstein, Runge, and
  Denzler]{requenamesa2021earthnet}
Christian Requena-Mesa, Vitus Benson, Markus Reichstein, Jakob Runge, and
  Joachim Denzler.
\newblock {EarthNet2021}: A large-scale dataset and challenge for earth surface
  forecasting as a guided video prediction task.
\newblock In \emph{CVPRW}, pages 1132--1142, 2021.

\bibitem[Ryan et~al.(2026)Ryan, Powell, Ling, and Wen]{ryan2026wetland}
Shawn Ryan, Megan Powell, Joanne Ling, and Li Wen.
\newblock Streamlining wetland vegetation mapping with {AlphaEarth} embeddings:
  Comparable accuracy to traditional methods with cleaner maps and minimal
  preprocessing.
\newblock \emph{Remote Sensing}, 18\penalty0 (2):\penalty0 293, 2026.

\bibitem[Salinas et~al.(2020)Salinas, Flunkert, Gasthaus, and
  Januschowski]{salinas2020deepar}
David Salinas, Valentin Flunkert, Jan Gasthaus, and Tim Januschowski.
\newblock {DeepAR}: Probabilistic forecasting with autoregressive recurrent
  networks.
\newblock \emph{International Journal of Forecasting}, 36\penalty0
  (3):\penalty0 1181--1191, 2020.

\bibitem[Sitzmann et~al.(2020)Sitzmann, Martel, Bergman, Lindell, and
  Wetzstein]{sitzmann2020siren}
Vincent Sitzmann, Julien N.~P. Martel, Alexander~W. Bergman, David~B. Lindell,
  and Gordon Wetzstein.
\newblock Implicit neural representations with periodic activation functions.
\newblock In \emph{Advances in Neural Information Processing Systems
  (NeurIPS)}, 2020.

\bibitem[Smith et~al.(2023)Smith, Fleming, and Geach]{smith2023earthpt}
Michael~J. Smith, Luke Fleming, and James~E. Geach.
\newblock {EarthPT}: A time series foundation model for {E}arth observation.
\newblock \emph{NeurIPS Climate Change AI Workshop}, 2023.

\bibitem[Sousa et~al.(2026)Sousa, Tebbutt, Jaffer, Young, Madhavapeddy, and
  Turner]{sousa2026downscaling}
Pedro Sousa, Will Tebbutt, Sadiq Jaffer, Robin Young, Anil Madhavapeddy, and
  Richard~E. Turner.
\newblock {E}arth observation embeddings are effective sub-grid descriptors for
  probabilistic weather downscaling.
\newblock \emph{arXiv preprint arXiv:2608.12271}, 2026.

\bibitem[Stewart et~al.(2026)Stewart, Fang, Corley, and
  Zhu]{stewart2026earthembeddings}
Adam~J. Stewart, Heng Fang, Isaac~A. Corley, and Xiao~Xiang Zhu.
\newblock {E}arth embeddings.
\newblock \emph{arXiv preprint arXiv:2608.03410}, 2026.

\bibitem[Sun et~al.(2021)Sun, Lei, Li, Tan, and Kuang]{sun2021scg}
Yuli Sun, Lin Lei, Xiao Li, Xiang Tan, and Gangyao Kuang.
\newblock Structure consistency-based graph for unsupervised change detection
  with homogeneous and heterogeneous remote sensing images.
\newblock \emph{IEEE Transactions on Geoscience and Remote Sensing},
  60:\penalty0 1--21, 2021.

\bibitem[Szwarcman et~al.(2024)Szwarcman, Roy, Fraccaro,
  et~al.]{szwarcman2024prithvi}
Daniela Szwarcman, Sujit Roy, Paolo Fraccaro, et~al.
\newblock {Prithvi-EO-2.0}: A versatile multi-temporal foundation model for
  earth observation applications.
\newblock \emph{arXiv preprint arXiv:2412.02732}, 2024.

\bibitem[Themistocleous et~al.(2020)Themistocleous, Papoutsa, Michaelides, and
  Hadjimitsis]{themistocleous2020pi}
Kyriacos Themistocleous, Christiana Papoutsa, Silas Michaelides, and Diofantos
  Hadjimitsis.
\newblock Investigating detection of floating plastic litter from space using
  {Sentinel-2} imagery.
\newblock \emph{Remote Sensing}, 12:\penalty0 2648, 2020.

\bibitem[Tseng et~al.(2025)Tseng, Fuller, Reil, Herzog, Beukema, Bastani,
  Green, Shelhamer, Kerner, and Rolnick]{tseng2025galileo}
Gabriel Tseng, Anthony Fuller, Marlena Reil, Henry Herzog, Patrick Beukema,
  Favyen Bastani, James~R. Green, Evan Shelhamer, Hannah Kerner, and David
  Rolnick.
\newblock {Galileo}: Learning global \& local features of many remote sensing
  modalities.
\newblock \emph{arXiv preprint arXiv:2502.09356}, 2025.

\bibitem[Verbesselt et~al.(2010)Verbesselt, Hyndman, Newnham, and
  Culvenor]{verbesselt2010bfast}
Jan Verbesselt, Rob Hyndman, Glenn Newnham, and Darius Culvenor.
\newblock Detecting trend and seasonal changes in satellite image time series.
\newblock \emph{Remote Sensing of Environment}, 114\penalty0 (1):\penalty0
  106--115, 2010.

\bibitem[Wang et~al.(2018)Wang, Xie, Zhang, Chen, Guo, Du, and
  Duan]{wang2018mbwi}
Xiaobiao Wang, Shunping Xie, Xueliang Zhang, Cheng Chen, Hao Guo, Jinkang Du,
  and Zheng Duan.
\newblock A robust multi-band water index ({MBWI}) for automated extraction of
  surface water from {Landsat 8 OLI} imagery.
\newblock \emph{International Journal of Applied Earth Observation and
  Geoinformation}, 68:\penalty0 73--91, 2018.

\bibitem[Wang et~al.(2023)Wang, Braham, Xiong, Liu, Albrecht, and
  Zhu]{wang2023ssl4eo}
Yi Wang, Nassim Ait~Ali Braham, Zhitong Xiong, Chenying Liu, Conrad~M.
  Albrecht, and Xiao~Xiang Zhu.
\newblock {SSL4EO-S12}: A large-scale multimodal, multitemporal dataset for
  self-supervised learning in {Earth} observation.
\newblock \emph{IEEE Geosci. Remote Sens. Mag.}, 11\penalty0 (3):\penalty0
  98--106, 2023.

\bibitem[Wei et~al.(2023)Wei, Jia, Wang, Cao, Qi, Zhao, and Yang]{wei2023pkad}
Haoran Wei, Kun Jia, Qiao Wang, Biao Cao, Jianbo Qi, Wenwu Zhao, and Jian Yang.
\newblock Real-time remote sensing detection framework of the earth's surface
  anomalies based on a priori knowledge base.
\newblock \emph{International Journal of Applied Earth Observation and
  Geoinformation}, 122, 2023.

\bibitem[Woo et~al.(2022)Woo, Liu, Sahoo, Kumar, and Hoi]{woo2022etsformer}
Gerald Woo, Chenghao Liu, Doyen Sahoo, Akshat Kumar, and Steven Hoi.
\newblock {ETSformer}: Exponential smoothing transformers for time-series
  forecasting.
\newblock \emph{arXiv preprint arXiv:2202.01381}, 2022.

\bibitem[Wu et~al.(2025)Wu, Mai, Lao, and Janowicz]{wu2025locdiff}
Zhangyu Wu, Gengchen Mai, Ni Lao, and Krzysztof Janowicz.
\newblock {LocDiff}: Identifying locations on {E}arth by diffusing in the
  {H}ilbert space.
\newblock In \emph{NeurIPS}, 2025.

\bibitem[Xia et~al.(2018)Xia, Chen, and Quan]{xia2018tai}
Haiping Xia, Yunhao Chen, and Jinling Quan.
\newblock A simple method based on the thermal anomaly index to detect
  industrial heat sources.
\newblock \emph{International Journal of Applied Earth Observation and
  Geoinformation}, 73:\penalty0 627--637, 2018.

\bibitem[Xiong et~al.(2024)Xiong, Wang, Zhang, Stewart, Hanna, Borth,
  Papoutsis, Saux, Camps-Valls, and Zhu]{xiong2024dofa}
Zhitong Xiong, Yi Wang, Fahong Zhang, Adam~J. Stewart, Jo{\"e}lle Hanna, Damian
  Borth, Ioannis Papoutsis, Bertrand~Le Saux, Gustau Camps-Valls, and
  Xiao~Xiang Zhu.
\newblock Neural plasticity-inspired multimodal foundation model for {E}arth
  observation.
\newblock \emph{arXiv preprint arXiv:2403.15356}, 2024.

\bibitem[Xu(2006)]{xu2006mndwi}
Hanqiu Xu.
\newblock Modification of normalised difference water index ({NDWI}) to enhance
  open water features in remotely sensed imagery.
\newblock \emph{International Journal of Remote Sensing}, 27:\penalty0
  3025--3033, 2006.

\bibitem[Yin et~al.(2019)Yin, Liu, Zhang, Wang, Shah, and
  Zimmermann]{yin2019gps2vec}
Yifang Yin, Zhenguang Liu, Ying Zhang, Sheng Wang, Rajiv~Ratn Shah, and Roger
  Zimmermann.
\newblock {GPS2Vec}: Towards generating worldwide {GPS} embeddings.
\newblock In \emph{ACM SIGSPATIAL}, 2019.

\bibitem[Zhang et~al.(2025)Zhang, Liang, Li, Ma, Xu, Ma, Xie, Li, Zhang, Tao,
  and Xia]{zhang2025units}
Yuxiang Zhang, Shunlin Liang, Wenyuan Li, Han Ma, Jianglei Xu, Yichuan Ma,
  Jiangwei Xie, Wei Li, Mengmeng Zhang, Ran Tao, and Xiang-Gen Xia.
\newblock {UniTS}: Unified spatio-temporal generative model for remote sensing.
\newblock \emph{arXiv preprint arXiv:2512.04461}, 2025.

\bibitem[Zhu and Woodcock(2014)]{zhu2014ccdc}
Zhe Zhu and Curtis~E. Woodcock.
\newblock Continuous change detection and classification of land cover using
  all available {Landsat} data.
\newblock \emph{Remote Sensing of Environment}, 144:\penalty0 152--171, 2014.

\bibitem[Zhu et~al.(2020)Zhu, Zhang, Yang, Aljaddani, Cohen, Qiu, and
  Zhou]{zhu2020cold}
Zhe Zhu, Junxue Zhang, Zhiqiang Yang, Amal~H. Aljaddani, Warren~B. Cohen, Shi
  Qiu, and Congliang Zhou.
\newblock Continuous monitoring of land disturbance based on {Landsat} time
  series.
\newblock \emph{Remote Sensing of Environment}, 238:\penalty0 111116, 2020.

\end{thebibliography}
}

\clearpage
\appendix
\section{Additional results}
\label{sec:appendix}

\subsection{Point vs.\ ensemble forecasts on S2\_Embed}
\label{sec:app_point_vs_ensemble}
\begin{figure*}[!ht]
  \centering
  \includegraphics[width=0.95\linewidth]{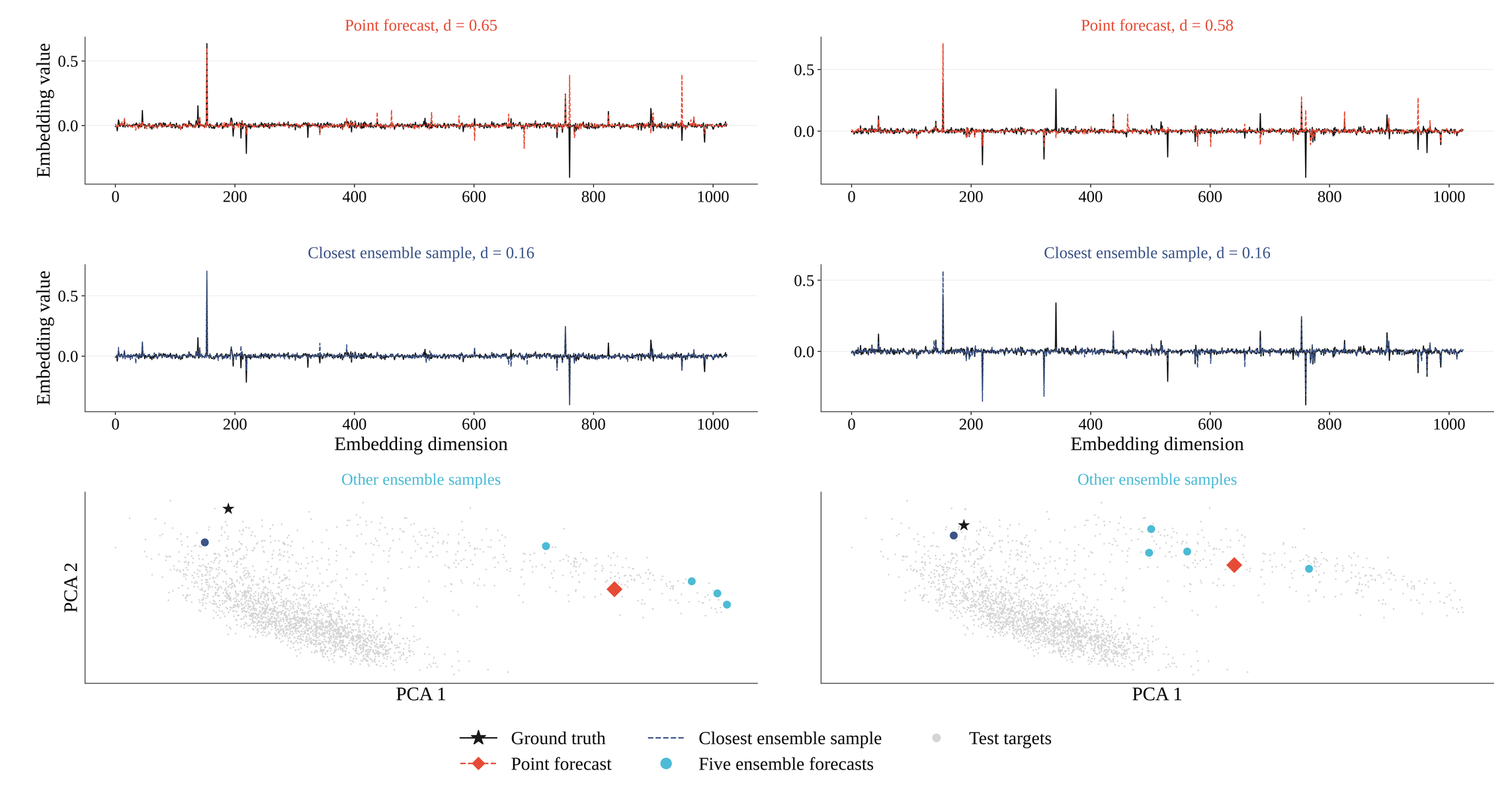}
  \caption{\textbf{Point \vs ensemble forecasts of \method{} on S2\_Embed.}
  Same layout as \cref{fig:point_vs_ensemble_s1}, over the 1024 dimensions.
  $d$ is the cosine distance to the ground truth.}
  \label{fig:point_vs_ensemble_s2}
\end{figure*}

\cref{fig:point_vs_ensemble_s2} repeats the visualisation of
\cref{fig:point_vs_ensemble_s1} for random test samples of S2\_Embed.

\subsection{Effect of the forecast lead time}
\label{sec:app_lead_time}
\begin{figure*}[t]
  \centering
  \includegraphics[width=\linewidth]{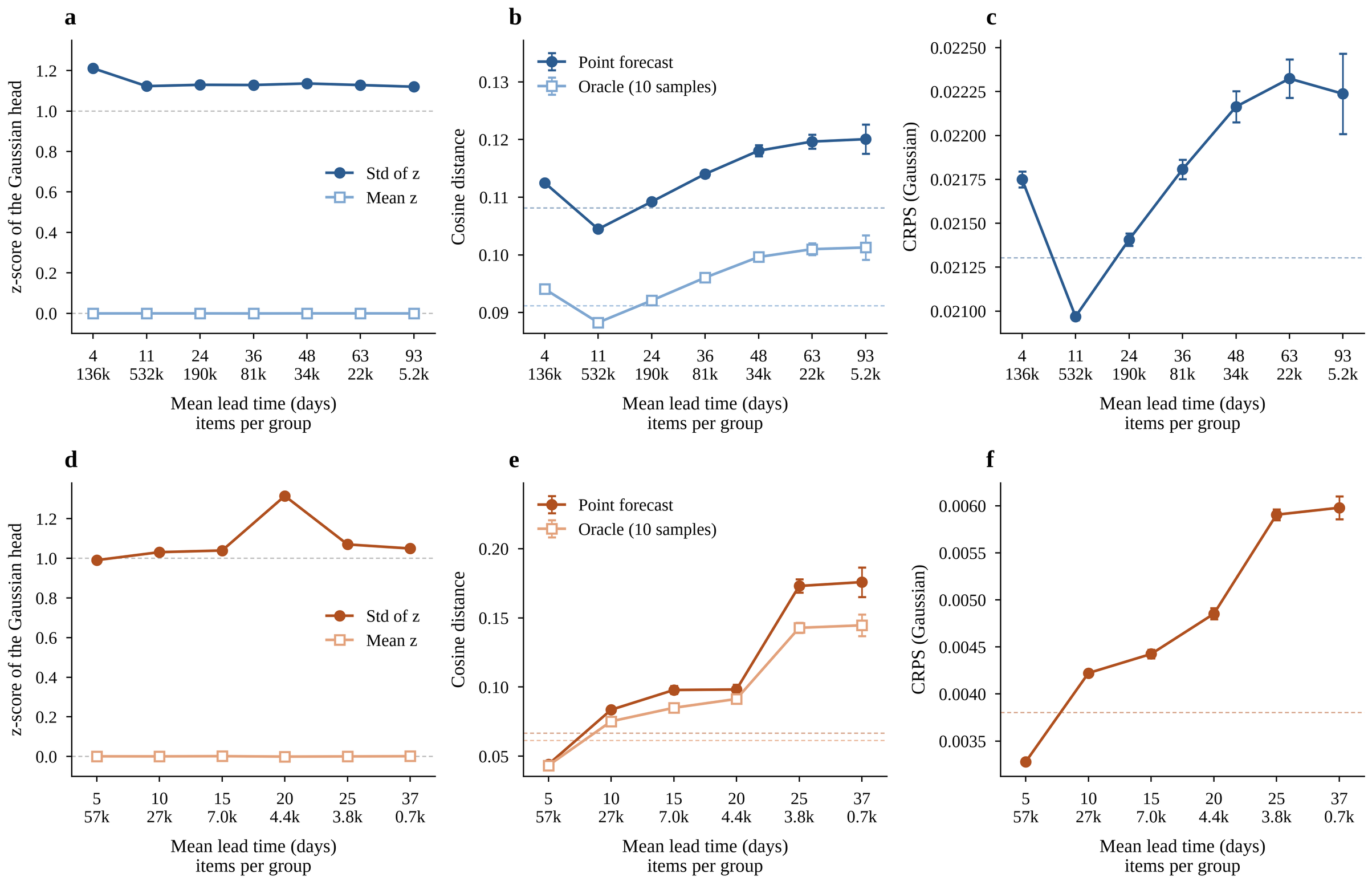}
  \caption{\textbf{Effect of the forecast lead time.} Test samples of S1\_Embed
  (a--c) and S2\_Embed (d--f) are grouped by the lead time, and the tick labels give the mean lead
  time and the number of items of each group. (a, d) Mean and std of the
  $z$-score of the Gaussian head; (b, e) cosine distance of the point forecast
  and of the best of ensemble members; (c, f) CRPS of the Gaussian head.
  Error bars are 95\% confidence intervals of the group mean, and dashed lines
  mark the value over the whole test set.}
  \label{fig:lead_time}
\end{figure*}

\cref{fig:lead_time} reports the metrics of \method{} per lead-time group on
S1\_Embed and S2\_Embed, as discussed in \cref{sec:exp_analysis}.

\subsection{Held-out predictions of the location-query INR}
\label{sec:app_qualitative}
\begin{figure*}[t]
  \centering
  \includegraphics[width=\linewidth]{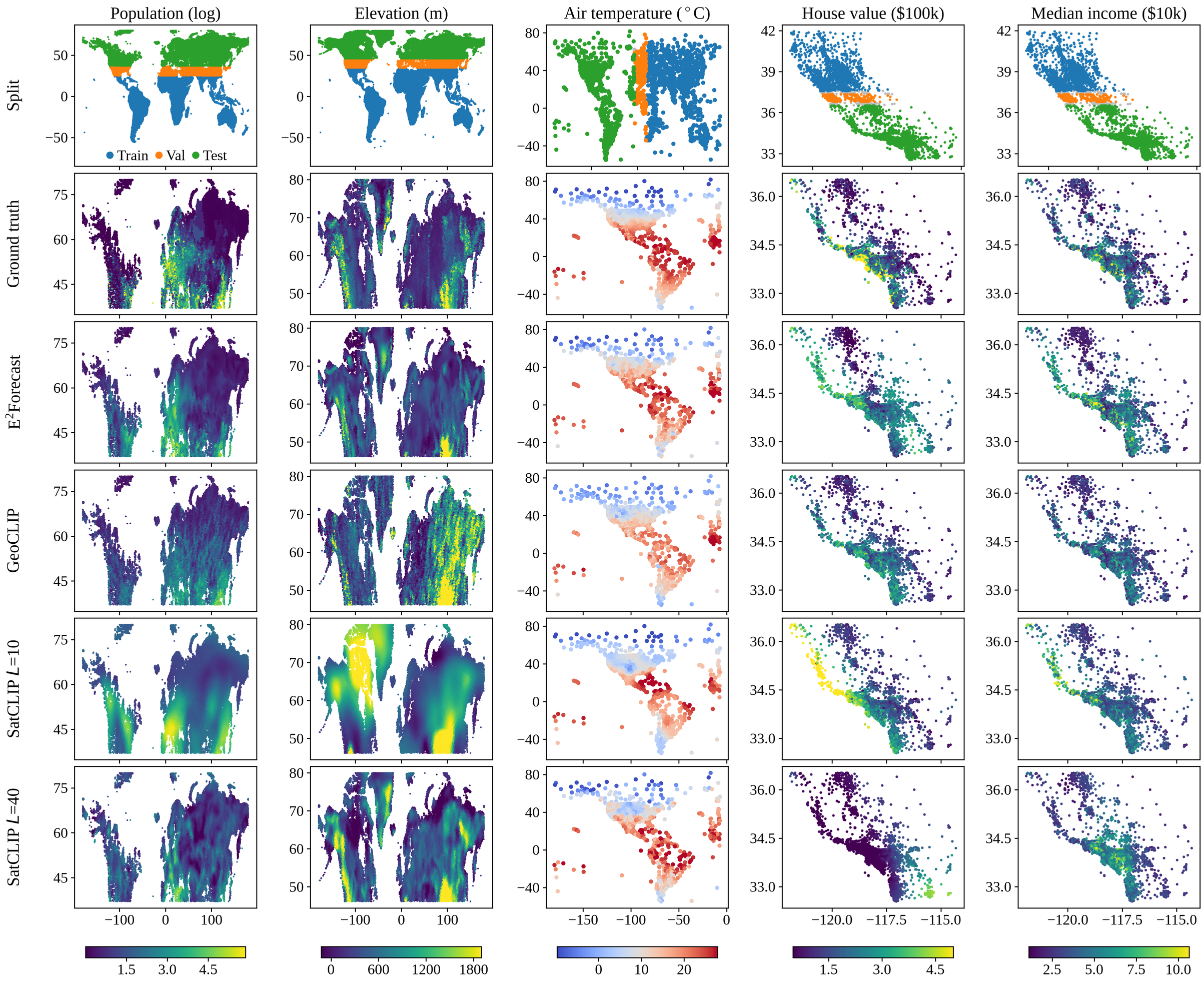}
  \caption{\textbf{Held-out predictions on all five tasks.} Top row: spatial
  split (\textcolor{blue}{train}, \textcolor{orange}{val},
  \textcolor{green!50!black}{test}). Below: ground truth and each method on the
  test band only, sharing the ground-truth colour scale per column; both SatCLIP
  rows use the ViT-16 tower. Axes are longitude and latitude in degrees.}
  \label{fig:qualitative}
\end{figure*}

\cref{fig:qualitative} shows the spatial splits and the held-out predictions of
each location encoder on the five regression tasks of \cref{sec:exp_spatial}.

\subsection{Anomaly change detection with different embedding spaces}
\label{sec:app_gsa_embedding}
\cref{tab:gsa_embedding} gives the average recall, precision and F$_1$ of
\method{} on GSA with each embedding space plotted in \cref{fig:gsa_embedding},
as discussed in \cref{sec:exp_analysis}.

\begin{table}[!h]
  \centering
  \small
  \setlength{\tabcolsep}{5pt}
  \begin{tabular}{@{}lccc@{}}
    \toprule
    Embedding space & R & P & F$_1$ \\
    \midrule
    OlmoEarth~\cite{herzog2025olmoearth}        & 65.48 & \underline{48.84} & \underline{48.70} \\
    \quad + Tessera~\cite{feng2026tessera}       & 65.61 & \textbf{49.65} & \textbf{49.29} \\
    \quad + AlphaEarth~\cite{brown2025alphaearth} & \underline{69.45} & 47.68 & 47.04 \\
    Galileo~\cite{tseng2025galileo}             & \textbf{70.54} & 45.58 & 46.53 \\
    CROMA~\cite{fuller2023croma}                & 69.25 & 34.25 & 36.82 \\
    SatMAE~\cite{cong2022satmae}                & 61.51 & 42.56 & 42.61 \\
    Clay~\cite{clay2024}                        & 60.77 & 45.89 & 43.64 \\
    \midrule
    ESIA~\cite{li2026esia} (best baseline)      & 54.99 & 40.05 & 41.58 \\
    \bottomrule
  \end{tabular}
  \caption{\textbf{Results of \method{} on GSA with different embedding spaces.}
  Average pixel-level recall (R), precision (P) and F$_1$ (\%) over the six
  event categories. ``+'' adds annual embeddings to OlmoEarth as extra
  conditions. Best and second best per column are in \textbf{bold} and
  \underline{underlined}.}
  \label{tab:gsa_embedding}
\end{table}

\end{document}